\documentclass[12pt]{article}

\usepackage{amsmath,amsthm, amsfonts, amssymb, amsxtra, amsopn}
\usepackage{pgfplots}
\pgfplotsset{compat=1.13}
\usepackage{pgfplotstable}
\usepackage{graphicx,grffile}
\usepackage{multirow}
\usepackage{booktabs}
\usepackage{listings}
\usepackage{cmap}
\usepackage{colortbl}
\usepackage{adjustbox}
\usepackage{epsfig}

\usepackage[tableposition=top,font=small,skip=5pt]{caption}
\usepackage{subcaption}
\usepackage{makecell} 

\usepackage[explicit]{titlesec}

\def\centroid#1{\mbox{centroid}(#1)}
\def\distortion#1{\mbox{distortion}_{#1}}

\PassOptionsToPackage{hyphens}{url}

\usepackage{hyperref}
\hypersetup{colorlinks=true,linkcolor=black,citecolor=black,urlcolor=blue,filecolor=black}
\hypersetup{pdfpagemode=UseNone,pdfstartview=}

\definecolor{darkgreen}{rgb}{0.125,0.5,0.169}

\usepackage{enumitem}
\setlist[itemize]{noitemsep, topsep=0pt}

\advance\oddsidemargin by -0.35in
\advance\textwidth by 0.7in

\advance\topmargin by -0.4in
\advance\textheight by 0.8in

\def\eref#1{(\ref{#1})}

\long\def\symbolfootnotetext[#1]#2{\begingroup%
\def\thefootnote{\fnsymbol{footnote}}\footnotetext[#1]{#2}\endgroup}

\newcommand\dunderline[3][-1pt]{{%
  \sbox0{#3}%
  \ooalign{\copy0\cr\rule[\dimexpr#1-#2\relax]{\wd0}{#2}}}}
\def\uuu{\kern-1pt\dunderline{0.75pt}{\phantom{M}}}

\DeclareMathOperator{\thth}{th}

\def\zz{\phantom{0}}

\title{Cluster Analysis and Concept Drift Detection in Malware}

\author{Aniket Mishra\footnotemark[1]\ \ \ 
Mark Stamp\footnotemark[1]\,\,\footnotemark[2]}

\begin{document}

\symbolfootnotetext[1]{Department of Computer Science, San Jose State University}
\symbolfootnotetext[2]{mark.stamp$@$sjsu.edu}

\maketitle

\abstract
Concept drift refers to gradual or sudden 
changes in the properties of data that affect the accuracy of machine learning models. 
In this paper, we address the problem of concept drift detection in the malware domain.
Specifically, we propose and analyze a clustering-based approach to detecting
concept drift. Using a subset of the KronoDroid dataset, malware samples are partitioned 
into temporal batches and analyzed using MiniBatch $K$-Means clustering. 
The silhouette coefficient is used 
as a metric to identify points in time where concept drift has likely occurred. To verify our
drift detection results, we train learning models under three realistic scenarios, 
which we refer to as static training, periodic retraining, and drift-aware retraining. 
In each scenario, we consider four supervised classifiers, namely, Multilayer Perceptron (MLP), 
Support Vector Machine (SVM), Random Forest, and XGBoost. 
Experimental results demonstrate that drift-aware retraining guided by 
silhouette coefficient thresholding 
achieves classification accuracy far superior to static models, and generally
within~1\%\ of periodic retraining, 
while also being far more efficient than periodic retraining. 
These results provide strong evidence that our clustering-based 
approach is effective at detecting concept drift, while also illustrating a 
highly practical and efficient fully automated approach to improved malware 
classification via concept drift detection.

\section{Introduction}\label{chap:intro}

In many applications of machine learning models, the underlying data is not static, 
and hence the features that a model was trained on might change over time. 
Such a change in features is generically referred to  
as concept drift. It is well known that malware writers modify existing malware for use 
in new attacks and to obfuscate malware in an attempt to evade various detection methods~\cite{Lout2024,luomaaho2023analysis}. 
This process of malware evolution can result in gradual or sudden changes in the 
features that machine learning models rely on to detect malware, thereby reducing 
the effectiveness of such models. Hence, concept drift detection is vitally important
in the malware domain, yet this topic appears to have received limited attention 
in the malware research literature. 

Traditionally, malware detection relied primarily on signature-based techniques.
Due to their inherently static nature, signatures
tend to fail under even relatively minor cases of concept drift~\cite{Alejandro}.
On the other hand,
machine learning based techniques, such as Hidden Markov Models (HMM), 
Support Vector Machines (SVM), 
Convolutional Neural Networks (CNN), and Random Forest classifiers, are generally
more robust~\cite{Anupama2022}. 
Consequently, machine learning techniques
have become the focus of malware detection research~\cite{Gibert}. 
However, these methods will also eventually fail 
once malware features sufficiently evolve beyond the training data.

It is well known that there are many obfuscation techniques
available that easily defeat signature scanning~\cite{bm},
and it is not difficult to adapt such techniques to 
defeat machine learning based techniques~\cite{Lin}.
Due to the widespread use of machine learning for malware detection,
and the ease with which malware can be modified---either for
the purpose of obfuscation or due to the 
inherent evolutionary nature of malware---concept drift is a
vitally important topic in the malware field. 

An obvious approach to concept drift detection is to statistically monitor features over time. 
However, it is important to automate the process of concept drift detection, and hence the use
of learning models to monitor features is a reasonable approach. Such an
approach has been previously considered, with HMM and SVM models used for concept 
drift detection in malware classification models~\cite{Sunhera,Lolitha,Mayuri}.

In this paper, we use a clustering based approach to detect concept drift.  
Specifically, we partition the sample of a malware family into
temporal subsets. We then analyze consecutive pairs of subsets using 
MiniBatch $K$-Means clustering~\cite{scikit-learn_mini_batch_kmeans}. 
The silhouette coefficient is used to quantify changes between consecutive clusterings, 
which enables us to detect points in time where significant concept drift has likely occurred.
We then verify that concept drift is being detected by considering various 
model retraining scenarios, which we now introduce.

We train various machine learning models, and use these models to 
classify the samples of a given family. To determine the effectiveness of our 
clustering-based concept drift detection strategy, we consider
the following three scenarios.
\begin{description}
\item[Static training]--- A model is trained on the initial subset of samples, 
and this model is used to classify all remaining
samples, without any retraining.
\item[Periodic retraining]--- A model is trained on the initial subset of samples, and the
model is retrained at regular periodic intervals. A trained model is
used for classification only until the next model is trained.
\item[Drift-aware retraining]--- This is similar to the periodic retraining scenario, 
except that retraining is 
triggered by concept drift detection, rather than regular time intervals.
\end{description}

For each of the three scenarios described above, we experiment with four classifiers. 
Specifically, under each scenario we train  
a Multilayer Perceptron (MLP) neural networks~\cite{scikit-learn_mlp_classifier}, 
Support Vector Machines (SVM)~\cite{scikit-learn_linear_svc}, 
Random Forests~\cite{scikit-learn_random_forest_classifier}, 
and XGBoost models~\cite{xgboost_sklearn_estimator}. We determine the 
accuracy of each of these models under each of the three retraining scenarios.
If we are accurately detecting concept drift, we expect that drift-aware 
retraining will yield strong results, while also being more efficient than
the periodic retraining scenario, in the sense of requiring significantly 
fewer models to be trained.

While many malware datasets are available, few have reliable timestamps associated with 
samples. Without such timestamps, any malware classification results are questionable, due to the
fact that future samples are almost certainly used to train models that are then used
to classify earlier samples, which is
generally not realistic. In any case, accurate timestamps are absolutely essential
for concept drift research. In this paper, we use the KronoDroid dataset~\cite{Krono}, 
since each sample in this dataset includes a timestamp,
and these timestamps appear to be accurate.

The remainder of this paper is organized as follows. In Section~\ref{chap:bg}, 
we discuss relevant background topics, including 
MiniBatch $K$-Means and silhouette coefficients, 
and we introduce each of the
four models that we use for classification.
Section~\ref{sect:rw} contains an overview of 
previous research into concept drift detection within the malware realm. 
Section~\ref{chap:imp} provides implementation details for our experiments. 
Section~\ref{chap:ER} presents our experimental results for 
each model across each of our three retraining scenarios. 
Finally, Section~\ref{chap:con} summarizes 
our findings and highlights potential directions for future research.

\section{Background}\label{chap:bg}

In this section, we discuss relevant background topics. First, we
introduce the topics of supervised and unsupervised learning. Then we 
discuss MiniBatch $K$-Means clustering in some detail, followed by
a brief introduction to each of the four learning techniques used in our experiments.
Finally, we discuss the silhouette coefficient in some detail, as it is 
fundamental to our concept drift detection strategy.

\subsection{Supervised and Unsupervised Learning}

Supervised learning refers to models that are trained on labeled datasets. 
Most popular classifiers are supervised techniques. 
Such methods have proven effective at detecting malware but, 
of course, all are adversely affected by concept drift. 
In this paper, we train various models as a way to test the effectiveness
of our proposed concept drift detection strategy. Specifically, the 
classification models that we consider
are MLP, SVM, Random Forest, and XGBoost, 
each of which is introduced below.

Unsupervised learning is based on unlabeled data. Techniques such as $K$-Means
and Expectation Maximization (EM) clustering are classic 
examples of unsupervised learning. Clustering is typically used
in a data exploration mode, where we are trying to analyze and understand data that
we know little about. This makes clustering a potentially useful approach 
for concept drift detection. In this paper, we experiment with MiniBatch $K$-Means 
as a tool for automatically detecting concept drift. Below, we introduce the 
MiniBatch $K$-Means algorithm, which is a version of $K$-Means that is more efficient
for large datasets.

\subsection{MiniBatch $K$-Means}

In this section, we first discuss the $K$-means algorithm. Then 
we consider the modification known as MiniBatch $K$-Means.

Suppose that we are given the~$n$ feature vectors~$X_1, X_2, \ldots, X_n$,
where each of the~$X_i$ is typically a vector of, say, $m$ real numbers. 
We assume that we want to partition these~$n$ feature vectors
into~$K$ clusters, where~$K$ is specified. 
We also assume that we have a distance function~$d(X_i,X_j)$
that is defined for all pairs of vectors.
Furthermore, we require that each vector~$X_i$ belongs to exactly one cluster.

We associate a centroid
with each cluster, where the centroid can be viewed as representative 
of its cluster---intuitively, a centroid is the center of mass of its cluster.
We denote the clusters as~$C_j$ with the 
corresponding centroid denoted as~$c_j$, for~$j=1,2,\ldots,K$.
Note that in $K$-means, the centroids need not belong to the set
of feature vectors.

Suppose that we have clustered our~$n$ vectors. 
Then we have a set of~$K$ centroids
$$
  c_1, c_2, c_3, \ldots, c_K , 
$$
where each vector in our dataset is associated with one centroid.
Let~$\centroid{X_i}$ denote the (unique) centroid of the cluster
to which~$X_i$ belongs. 
Then the centroids determine the clusters, in the sense that 
$$
  c_j = \centroid{X_i} 
$$
implies~$X_i$ belongs to cluster~$C_j$. 

Before we can cluster data based on the outline above, 
we need to answer the following two basic questions.
\begin{enumerate}
\item How do we determine the centroids~$c_j$?
\item How do we determine the clusters? That is,
we need to specify the function~$\centroid{X_i}$
that assigns data points to centroids.
\end{enumerate}
At the very least, we need a method to compare clusterings, that is,
we must have a way to determine whether one clustering is better than another.
The $K$-Means algorithm uses a simple method for
quantifying cluster quality. 

Intuitively, the more compact a cluster, the better. Of course, this will
depend on the vectors~$X_i$ and the number of clusters~$K$. Since the data is given,
and we assume that~$K$ has been specified, 
we have no control over the~$X_i$ or~$K$.  But, we do have
control over the centroids~$c_j$ and the assignment of points to centroids via the
function~$\centroid{X_i}$. The choice of centroids and the assignment of
points to centroids will clearly influence the compactness
(or ``shape'') of the resulting clusters. 

We define
\begin{equation}\label{eq:cluster_distortion}
  \distortion{} = \sum_{i=1}^n d\bigl(X_i,\centroid{X_i}\bigr) .
\end{equation}
The smaller the~$\distortion{}$, the better, since a smaller~$\distortion{}$
implies that individual clusters are more compact.

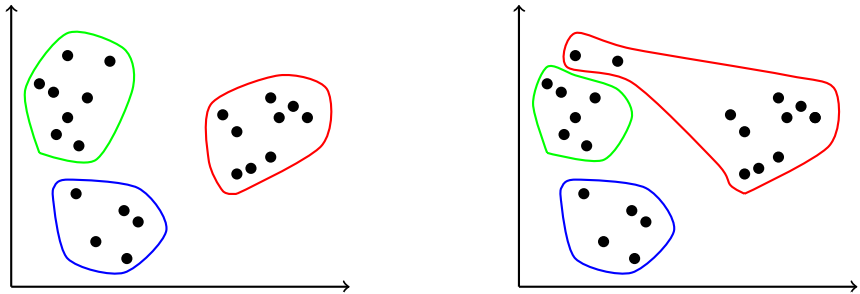
\begin{figure}[!htb]
\centering
        \begin{tikzpicture}[scale=0.75]
    %
    %
    \draw[thick,color=black,fill=black] (0.8,2.7) circle (0.08);
    \draw[thick,color=black,fill=black] (1.0,3.0) circle (0.08);
    \draw[thick,color=black,fill=black] (1.2,2.5) circle (0.08);
    \draw[thick,color=black,fill=black] (1.35,3.35) circle (0.08);
    \draw[thick,color=black,fill=black] (0.75,3.45) circle (0.08);
    \draw[thick,color=black,fill=black] (1.75,4.0) circle (0.08);
    \draw[thick,color=black,fill=black] (0.5,3.6) circle (0.08);
    \draw[thick,color=black,fill=black] (1.0,4.1) circle (0.08);
    
    \draw[thick,color=black,fill=black] (2.0,1.35) circle (0.08);
    \draw[thick,color=black,fill=black] (1.5,0.8) circle (0.08);
    \draw[thick,color=black,fill=black] (1.15,1.65) circle (0.08);
    \draw[thick,color=black,fill=black] (2.25,1.15) circle (0.08);
    \draw[thick,color=black,fill=black] (2.05,0.5) circle (0.08);

    \draw[thick,color=black,fill=black] (4.0,2.0) circle (0.08);
    \draw[thick,color=black,fill=black] (4.25,2.1) circle (0.08);
    \draw[thick,color=black,fill=black] (4.6,2.3) circle (0.08);
    \draw[thick,color=black,fill=black] (5.25,3.0) circle (0.08);
    \draw[thick,color=black,fill=black] (4.75,3.0) circle (0.08);
    \draw[thick,color=black,fill=black] (4.0,2.75) circle (0.08);
    \draw[thick,color=black,fill=black] (3.75,3.05) circle (0.08);
    \draw[thick,color=black,fill=black] (4.6,3.35) circle (0.08);
    \draw[thick,color=black,fill=black] (5.0,3.2) circle (0.08);
    
    \draw[blue,thick] plot [smooth] coordinates {
    (0.75,1.5) (1.0,0.5) (2.0,0.25) (2.75,1.0) (2.25,1.75) (1,1.9) (0.75,1.75) (0.75,1.5)};
    \draw[red,thick] plot [smooth] coordinates {
    (4.0,1.65) (5.5,2.5) (5.6,3.5) (4.75,3.75) (3.55,3.25) (3.5,2.25) (3.75,1.7) (4.0,1.65)};
    \draw[green,thick] plot [smooth] coordinates {
    (0.5,2.375) (1.5,2.25) (2.15,3.5) (2.0,4.22) (1.0,4.5) (0.25,3.5) (0.5,2.375)};
        
        
     \draw[thick,color=black,->] (0,0) -- (6,0); 
     \draw[thick,color=black,->] (0,0) -- (0,5); 
     \draw[thick,color=white,->] (0,-0.2) -- (6,-0.2); 
     \draw[thick,color=white,->] (-0.2,0) -- (-0.2,5); 

    %
    %
    \draw[thick,color=black,fill=black] (9.8,2.7) circle (0.08);
    \draw[thick,color=black,fill=black] (10.0,3.0) circle (0.08);
    \draw[thick,color=black,fill=black] (10.2,2.5) circle (0.08);
    \draw[thick,color=black,fill=black] (10.35,3.35) circle (0.08);
    \draw[thick,color=black,fill=black] (9.75,3.45) circle (0.08);
    \draw[thick,color=black,fill=black] (9.5,3.6) circle (0.08);
    
    \draw[thick,color=black,fill=black] (11.0,1.35) circle (0.08);
    \draw[thick,color=black,fill=black] (10.5,0.8) circle (0.08);
    \draw[thick,color=black,fill=black](10.15,1.65) circle (0.08);
    \draw[thick,color=black,fill=black] (11.25,1.15) circle (0.08);
    \draw[thick,color=black,fill=black] (11.05,0.5) circle (0.08);

    \draw[thick,color=black,fill=black] (13.0,2.0) circle (0.08);
    \draw[thick,color=black,fill=black] (13.25,2.1) circle (0.08);
    \draw[thick,color=black,fill=black] (13.6,2.3) circle (0.08);
    \draw[thick,color=black,fill=black] (14.25,3.0) circle (0.08);
    \draw[thick,color=black,fill=black] (14.25,3.0) circle (0.08);
    \draw[thick,color=black,fill=black] (13.75,3.0) circle (0.08);
    \draw[thick,color=black,fill=black] (13.0,2.75) circle (0.08);
    \draw[thick,color=black,fill=black] (12.75,3.05) circle (0.08);
    \draw[thick,color=black,fill=black] (13.6,3.35) circle (0.08);
    \draw[thick,color=black,fill=black] (14.0,3.2) circle (0.08);

    \draw[thick,color=black,fill=black] (10.75,4.0) circle (0.08);
    \draw[thick,color=black,fill=black] (10.0,4.1) circle (0.08);

    \draw[blue,thick] plot [smooth] coordinates {
    (9.75,1.5) (10.0,0.5) (11.0,0.25) (11.75,1.0) (11.25,1.75) (10,1.9) (9.75,1.75) (9.75,1.5)};
    \draw[red,thick] plot [smooth] coordinates {
    (13.0,1.65) (14.5,2.5) (14.6,3.5) (13.75,3.75) (11.0,4.22) (10.0,4.5)  (9.85,3.9) (11,3.625)
    (12.5,2.2) (12.75,1.8) (13.0,1.65)}; 
    \draw[green,thick] plot [smooth] coordinates {
    (9.5,2.375) (10.5,2.25) (11.0,3.0) (10.75,3.5) (10.0,3.75) (9.5,3.9) (9.25,3.25) (9.5,2.375)};


     \draw[thick,color=black,->] (9,0) -- (15,0); 
     \draw[thick,color=black,->] (9,0) -- (9,5); 
     \draw[thick,color=white,->] (9,-0.2) -- (15,-0.2); 
     \draw[thick,color=white,->] (8.8,0) -- (8.8,5); 
   
    \end{tikzpicture}
\caption{Less distortion and more distortion}\label{fig:clustDistortion}
\end{figure}

For example, consider the data in Figure~\ref{fig:clustDistortion}, where the same
data points are clustered in two different ways. It is clear that the clustering on 
the left-hand side in Figure~\ref{fig:clustDistortion} has a smaller $\distortion{}$ than 
that on the right-hand side. Therefore, we would say that the left-hand clustering is superior,
at least with respect to the measure of~$\distortion{}$.

Suppose that we want to minimize the $\distortion{}$. 
First, we note that the $\distortion{}$ depends on~$K$, 
since more clusters means more centroids and, all else being equal, 
the larger the value of~$K$, the closer each point will tend to
be to its centroid. Hence, we write $\distortion{K}$ to emphasize the 
dependence on the hyperparameter~$K$. As mentioned above, 
we assume that~$K$ is specified in advance.

The problem we want to solve can be stated precisely as follows.
\begin{equation}\label{eq:clustProb}
\begin{split}
  & \mbox{Given: } K \mbox{ and data points } X_1,X_2,\ldots, X_n\\[-0.75ex]
  & \mbox{Minimize: } \distortion{K} =  \sum_{i=1}^n d\bigl(X_i,\centroid{X_i}\bigr) .
\end{split}
\end{equation}
Finding an exact solution to this problem is computationally infeasible,
but we can derive a very simple, iterative approximation.

We claim that a solution to~\eref{eq:clustProb} must satisfy the following two conditions.
\begin{description}
\item[Condition~1]--- Each~$X_i$ is clustered according to its nearest centroid,
i.e., if~$X_i$ belongs to cluster~$C_j$, 
then~$d(X_i,c_j) \leq d(X_i,c_{\ell})$ for all~$\ell\in\{1,2,\ldots,K\}$,
where the~$c_{\ell}$ are the centroids.
\item[Condition~2]--- Each centroid is located at the center of its cluster.
\end{description}

To verify the necessity of Condition~1,
suppose that~$X_i$ is in cluster~$C_j$ and that~$d(X_i,c_{\ell}) < d(X_i,c_{j})$ 
for some~$\ell$. Then by simply reassigning~$X_i$ to cluster~$\ell$, we 
will reduce~$\distortion{K}$.
Condition~2  is only slightly challenging to prove.
In any case, these two conditions suggest an approximation 
algorithm.

From Condition~1, we see that given any clustering for which there are 
points that are not clustered based on
their nearest centroid, we can improve the clustering by simply reassigning
such data points to their nearest centroid. By Condition~2,
we always want the centroids to be at the center of the clusters. So, given any
clustering, we may improve it---and it cannot get worse---by performing 
either of the following two steps.

\begin{description}
\item[Step~1]--- Assign each data point to its nearest centroid.
\item[Step~2]--- Recompute the centroids so that each lies at the center of
its respective cluster.
\end{description}

It is clear that no improvement will result from applying 
Step~1 more than once in succession, and the same holds true for Step~2.
However, by alternating between these two steps, we obtain an iterative process that
yields a series of solutions that will generally tend to improve, and even in the worst
case, the solution cannot get worse. This is the $K$-means algorithm.

The $K$-Means algorithm is a hill climb, and hence we
are only assured of finding local maximum and, as with any hill climb,
the maximum we obtain will depend on our choice for the initial conditions. 
For $K$-means, the initial conditions correspond to the initial selection of centroids.
Therefore, it can be beneficial to repeat the algorithm multiple times with
different initializations of the centroids.

In the MiniBatch $K$-Means algorithm, instead of using the entire dataset to update the centroids, only
a small subset of the vectors are considered at each iteration. The algorithm 
can be stated as follows~\cite{minibatch_k_means}.
\begin{enumerate}
\item Centroids are initialized at random
\item As in standard $K$-Means, each data point is assigned to its nearest centroid
\item A minibatch of data is randomly selected from the dataset
\item The centroids are recomputed using only the points from the minibatch
\item If there is sufficient change in the centroids, goto~2
\end{enumerate}
MiniBatch $K$-Means is claimed to be more efficient---both in the sense of faster convergence
and requiring less memory---as compared to standard $K$-Means.
For the experiments discussed in this paper, we use the MiniBatch $K$-Means implementation 
at~\cite{scikit-learn_mini_batch_kmeans}.

\subsection{Classification Models}

For all of our classification results, we experiment with four distinct learning algorithms. As mentioned above, these 
four classification algorithms are MLP, SVM, Random Forest, and XGBoost. In this section,
we briefly introduce each of these algorithms.

\subsubsection{MLP}

Multilayer Perceptrons (MLP) are a fundamental class of feedforward artificial neural networks. 
MLPs are capable of learning complex, nonlinear relationships between the input and output spaces. 
For our MLP experiments, we use the implementation at~\cite{scikit-learn_mlp_classifier}.
In all of our MLP models, we use the popular ReLU activation functions, and
we also use the Adam optimizer to adjust the learning rate during training. 
The hyperparameters of an MLP include the number of neurons, the number of layers, 
and the type of regularization. 

As mentioned, one of the advantages of MLPs is their ability to model complex relationships. 
However, the tradeoff is that larger amounts of data and extended training times may be required. 
Batch normalization (BatchNorm) can be used to speed convergence and stabilize the training process. 
Regularization techniques are important for optimizing performance and to mitigate 
issues such as vanishing and exploding gradients~\cite{towardsdatascience_mlp}.

\subsubsection{SVM}

Support Vector Machine (SVM) is a classifier that attempts to find a separating hyperplane, where
the hyperplane maximize the margin (i.e., separation) between the two classes in the feature space. 
For all of our SVM models, we use the Support Vector Classifier implementation
at~\cite{scikit-learn_linear_svc}.

A key hyperparameter of an SVM is
the regularization parameter~$C$. This parameter balances the margin and classification accuracy.
A higher value for~$C$ results in a tighter margin, but can result in overfitting,
whereas a smaller~$C$ value increases the margin of separation, but can lead to 
underfitting~\cite{ibm_svm}.

\subsubsection{Random Forest}

Random Forest is an ensemble learning algorithm that consists of multiple decision trees. 
A random subset of data and features are used to build each tree using a process known as bagging. 
This approach helps reduce the likelihood of overfitting and also reduces the variance. During the classification 
stage, a majority vote among the component decision trees is used.
Random Forest models are considered to be particularly strong in cases with noisy data,
and the can handle both numerical and categorical variables. Another advantage is 
that a Random Forest provides details on feature importance~\cite{ibm_random_forest}.

Random Forest hyperparameters include the number of trees and the maximum depth of each tree. 
As the number of trees are increased, the stability increases and the variance is reduced. 
However, more trees generally leads to longer training times. Random Forests generally perform
well, but the technique may struggle on data with highly correlated features, as this tends to
result in redundant trees. To address this issue, dimensionality reduction techniques can be 
applied to the data before training a Random Forest model.

\subsubsection{XGBoost}

XGBoost (eXtreme Gradient Boosting) is a very efficient version of gradient boosting. 
As with Random Forests, XGBoost models are based on decisions trees.
However, unlike a Random Forest, an XGBoost model is determined sequentially,
in the sense that each tree focuses on reducing the residual errors of the previous tree. 
Overfitting is reduced by minimizing a differentiable loss function, and the technique 
also employs various regularization strategies. In XGBoost, regularization serves to control model 
complexity which, in turn, helps reduce overfitting.  XGBoost is also noted for its
ability to effectively deal with high dimensional data. 

Since XGBoost is based on decision trees many of the hyperparameters are similar to those
of a Random Forest. Some of the important hyperparameters in XGBoost include 
the learning rate, the maximum tree depth, and the number of estimators~\cite{ibm_xgboost}. 

\subsection{Silhouette Coefficient}

When clustering data, intuitively,
we want each cluster to be cohesive,
in the sense that the points within a cluster are close together. 
In addition, we would like the points in different 
clusters to be well separated.
That is, the more cohesive the individual clusters and
the more separation between the clusters, the better. 
In this section, we first provide formulae for
explicitly computing cohesion and separation, relative to a given data point. 
Then we combine these quantities into a single number to obtain the so-called 
silhouette coefficient, relative to the given data point. 
Finally, we argue that the average silhouette coefficient
over all data points provides a sensible measure of cluster quality.

Suppose that we have a set of~$n$ feature vectors~$X_1,X_2,\ldots,X_n$
partitioned into~$K$ clusters, $C_1,C_2,\ldots,C_K$. Let~$n_i$ be the
number of elements in cluster~$C_i$. Note that~$n_1+n_2+\cdots+n_K = n$.

Select a feature vector~$X_i$. Then~$X_i\in C_j$ for some~$j\in\{1,2,\ldots,K\}$.
Let~$a$ be the average distance from~$X_i$ to all other points in 
its cluster~$C_j$. That is, 
\begin{equation}\label{eq:clusterAAA}
  a =  \frac{1}{n_j - 1}\sum_{\substack{Y\in C_j\\ Y\neq X_i}} d(X_i,Y) .
\end{equation}
In the degenerate case where~$n_j = 1$, let~$a=0$.
Also, for each cluster~$C_{\ell}$, such that~$\ell\neq j$,  
let~$b_{\ell}$ be the average distance from~$X_i$
to the other vectors in cluster~$C_{\ell}$, that is,
$$
  b_{\ell} = \frac{1}{n_{\ell}}\sum_{Y\in C_{\ell}} d(X_i,Y) .
$$
Finally, we define~$b$ to be the smallest of the~$b_{\ell}$, that is,
\begin{equation}\label{eq:clusterBBB}
  b = \!\!\!\!\min_{\substack{\ell\in\{1,2,\ldots,K\}\\ \ell\neq j}}\!\!\!\! b_{\ell} .
\end{equation}
The value~$a$ in~\eref{eq:clusterAAA} is a measure of the cohesion of cluster~$C_j$,
relative to the point~$X_i$, while~$b$ in~\eref{eq:clusterBBB} provides 
a measure of separation relative to the vector~$X_i$.

We define the  silhouette coefficient of feature vector~$X_i$ as 
$$
  S(X_i) = \frac{b-a}{\max(a,b)} .
$$
For any reasonable clustering, we expect that~$b > a$, in which case
$$
  S(X_i) = 1 - \frac{a}{b} .
$$

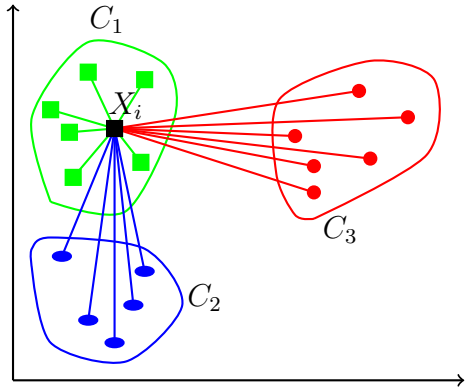
\begin{figure}[!htb]
\centering
        \begin{tikzpicture}[scale=1.0]

    \draw[thick,color=green,fill=green] (0.7,2.6) rectangle (0.9,2.8);
    \draw[thick,color=green,fill=green] (1.6,2.8) rectangle (1.8,3.0);
    \draw[thick,color=green,fill=green] (0.65,3.2) rectangle (0.85,3.4);
    \draw[thick,color=green,fill=green] (1.65,3.9) rectangle (1.85,4.1);
    \draw[thick,color=green,fill=green] (0.4,3.5) rectangle (0.6,3.7);
    \draw[thick,color=green,fill=green] (0.9,4.0) rectangle (1.1,4.2);

    \draw[thick,color=blue,fill=blue] (1.75,1.45) ellipse (0.12 and 0.06);
    \draw[thick,color=blue,fill=blue] (1.0,0.8) ellipse (0.12 and 0.06);
    \draw[thick,color=blue,fill=blue] (0.65,1.65) ellipse (0.12 and 0.06);
    \draw[thick,color=blue,fill=blue] (1.6,1.0) ellipse (0.12 and 0.06);
    \draw[thick,color=blue,fill=blue] (1.35,0.5) ellipse (0.12 and 0.06);

    \draw[thick,color=red,fill=red] (4.0,2.5) circle (0.08);
    \draw[thick,color=red,fill=red] (4.75,2.95) circle (0.08);
    \draw[thick,color=red,fill=red] (5.25,3.5) circle (0.08);
    \draw[thick,color=red,fill=red] (4.0,2.85) circle (0.08);
    \draw[thick,color=red,fill=red] (3.75,3.25) circle (0.08);
    \draw[thick,color=red,fill=red] (4.6,3.85) circle (0.08);

    \draw[blue,thick] plot [smooth] coordinates {
    (0.25,1.5) (0.5,0.5) (1.5,0.25) (2.25,1.0) (1.75,1.75) (0.5,1.9) (0.25,1.75) (0.25,1.5)};
    \draw[red,thick] plot [smooth] coordinates {
    (4.0,2.15) (5.5,3.0) (5.6,4.0) (4.75,4.25) (3.55,3.75) (3.5,2.75) (3.75,2.2) (4.0,2.15)};
    \draw[green,thick] plot [smooth] coordinates {
    (0.5,2.375) (1.5,2.25) (2.15,3.5) (2.0,4.22) (1.0,4.5) (0.25,3.5) (0.5,2.375)};
        
    \draw[thick,color=green] (1.35,3.35) -- (0.8,2.7);
    \draw[thick,color=green] (1.35,3.35) -- (1.7,2.9);
    \draw[thick,color=green] (1.35,3.35) -- (0.75,3.3);
    \draw[thick,color=green] (1.35,3.35) -- (1.75,4.0);
    \draw[thick,color=green] (1.35,3.35) -- (0.5,3.6);
    \draw[thick,color=green] (1.35,3.35) -- (1.0,4.1);

    \draw[thick,color=blue] (1.35,3.35) -- (1.75,1.45);
    \draw[thick,color=blue] (1.35,3.35) -- (1.0,0.8);
    \draw[thick,color=blue] (1.35,3.35) -- (0.65,1.65);
    \draw[thick,color=blue] (1.35,3.35) -- (1.6,1.0);
    \draw[thick,color=blue] (1.35,3.35) -- (1.35,0.5);

    \draw[thick,color=red] (1.35,3.35) -- (4.0,2.5);
    \draw[thick,color=red] (1.35,3.35) -- (4.75,2.95);
    \draw[thick,color=red] (1.35,3.35) -- (5.25,3.5);
    \draw[thick,color=red] (1.35,3.35) -- (4.0,2.85);
    \draw[thick,color=red] (1.35,3.35) -- (3.75,3.25);
    \draw[thick,color=red] (1.35,3.35) -- (4.6,3.85);

    \draw[thick,color=black,fill=black] (1.25,3.25) rectangle (1.45,3.45);

        
    \draw[thick,color=black,->] (0,0) -- (6,0); 
    \draw[thick,color=black,->] (0,0) -- (0,5); 
   
    \node at (1.5,3.65){$X_i$};
    \node at (1.25,4.8){$C_1$};
    \node at (2.55,1.1){$C_2$};
    \node at (4.35,2.0){$C_3$};

    \end{tikzpicture}
\caption{Silhouette coefficient example}\label{fig:clustSilEx}
\end{figure}

For example, consider the clusters in Figure~\ref{fig:clustSilEx}. 
In this case, $a$ is the average length of the lines connecting~$X_i$
to the other vectors (i.e., the squares) in its cluster, $C_1$. Also, $b_2$ is the average 
length of the lines from~$X_i$ to the vectors (i.e., ovals) in~$C_2$,
and~$b_3$ is the average length of the lines 
from~$X_i$ to the vectors (i.e., circles) in~$C_3$.
Letting~$b=\min\{b_2,b_3\}$, the silhouette coefficient 
for~$x_i$ is given by~$S(x_i)=1-a/b$.

If~$X_i$ is much closer to the vectors in its own cluster, as compared to
its distance to any other cluster, then~$a$ is much less than~$b$, 
and~$S(X_i)\approx 1$.
On the other hand, if~$X_i$ is relatively far from vectors in its own cluster and
close to vectors in at least one other cluster, then~$S(X_i)$ will be close to~0.
Therefore, the larger the silhouette coefficient~$S(X_i)$, the better. 
The average silhouette coefficient 
$$
  s = \frac{1}{n} \sum_{i=1}^n S(X_i)
$$
provides an intrinsic measure of the overall quality of a given clustering.
The bottom line is that the average silhouette coefficient combines 
cohesion and separation into a single number that provides
a useful intrinsic measure of cluster quality. 

\subsection {Silhouette Coefficient and Concept Drift}

For our purposes, 
we expect that if no concept drift has occurred, a consecutive pair of
temporal subsets will cluster similarly to the previous pair, yielding a small
difference in the average silhouette coefficient. In contrast, when concept
drift has occurred between consecutive temporal subsets,
we will see a more substantial difference in the average
silhouette coefficient, as compared to the clustering of the previous pair. 
Thus, we propose to use the 
average silhouette coefficient as a way to detect changes 
in feature vectors, which implies concept drift.

Recall from Section~\ref{chap:intro} that we consider three different scenarios, 
which we denote as static training, periodic retraining, and drift-aware retraining.
In the static training scenario, a model is trained on the initial temporal subset
and no retraining occurs; in the periodic retraining scenario, 
 a model is retrained at regular intervals without regard to concept drift detection;
while in the drift-aware retraining scenario, 
the model is retrained only when significant concept drift is 
detected via changes in silhouette coefficient. Note that only in the 
drift-aware scenario are we using the silhouette coefficient results.

\section{Related Work}\label{sect:rw}

In this section, our focus is to provide a selective survey of previous work in concept drift detection 
within the malware domain. While the topic of concept drift in malware has been considered in the 
literature, given its importance, it is surprising that more research has not been conducted.

Concept drift is relevant in many areas of cybersecurity~\cite{9027485}, but none more so than 
malware, where detection systems must adapt to ongoing changes. In the malware context,
concept drift can occur any time that new malware samples are created from existing families. 
Modifications are made to existing malware to implement new attacks, or for the 
purpose of obfuscation~\cite{10.1145/2381896.2381910}. Regardless of the reason for modifications,
concept drift will likely occur whenever samples of a malware family are 
modified to any significant degree. 

In recent years, the importance of detecting concept drift has been recognized. In the cybersecurity field, 
concept drift has been studied in areas as diverse as spam filtering and credit card fraud detection~\cite{Jog}. 
However, malware detection poses unique challenges, due to strategies that can be used to actively 
evade detection. Techniques such as polymorphism and metamorphism enable malware developers
to easily modify the features of existing malware families~\cite{Sharma_2014}. Such 
readily available obfuscation techniques further emphasize the need for a robust and automated 
concept drift detection process. It is also important to note that model training is generally
costly, and hence we want to minimize the frequency with which models are retrained.
Thus, detecting concept drift and retraining only when necessary is desirable.

In the interesting paper~\cite{Molina}, the authors detect concept drift in Android 
malware by monitoring the performance of a trained model, using techniques pioneered 
in~\cite{Page} and~\cite{Hinckley}. By retraining only when drift is detected,
they obtain detection results that are essentially equivalent to periodic retraining,
but the process is far more efficient, since fewer models need to be trained.

The research in~\cite{Alejandro} is primarily focused on the validity of timestamps in
the context of concept drift detection for Android malware. The authors develop 
an ``internal'' timestamp that appears to be more accurate than 
typical timestamps.

The paper~\cite{Karbab} relies on Natural Language Processing (NLP)
and machine learning techniques to adapt to changes in malware families. 
The goal is to cluster samples into their respective families.

In~\cite{Xu}, the authors develop DroidEvolver which includes a method 
for automatically updating an Android malware
detection model without retraining. The approach is highly efficient 
and the authors claim that it only reduces the performance
slightly over an extended period of time, as compared to state-of-the-art methods.

The research in~\cite{Kan} is focused on perceived flaws in DroidEvolver, which cause models
to perform much worse than claimed in~\cite{Xu}. The authors of~\cite{Kan} develop and test
DroidEvolver\texttt{++}, which they claim addresses the issues discovered in DroidEvolver.

The research in~\cite{Chen} focuses on a supposed distinction between 
feature-space drift (defined as new features introduced by new 
samples) and data-space drift (defined as data distribution 
shift over existing features) in the malware context. The authors 
claim that data-space drift consistently dominates, and they consider
this to be surprising.

The authors of~\cite{Yizheng} employ a hierarchical contrastive learning approach to update 
malware detection models. Their results show significant improvement over more costly 
retraining techniques.

The work presented in the remainder of this paper
is---in terms of the structure of our experiments---most similar to that in the paper~\cite{Molina}.
However, the actual techniques that we employ for concept drift detection are 
most closely related to those in the series of papers~\cite{Sunhera,Lolitha,Mayuri}
and, in fact, this paper can be viewed as a continuation and extension this series of papers.

\section{Implementation}\label{chap:imp}

This section includes a discussion of the dataset, the features, and preprocessing of the data.
We then discuss the three training scenarios in greater precision, and other relevant implementation 
details are given.

\subsection{Dataset}

This research utilizes the KronoDroid dataset, which contains~41,382 Android malware samples
that are categorized into~240 distinct families~\cite{Krono}. The dataset includes metadata that enables 
us to segment each family into temporal batches. This information is essential for the study 
of malware evolution and concept drift. Each sample in the KronoDroid dataset includes a variety
of extracted features, including permissions, API calls, manifest data, and various behavioral signatures.
In short, the KronoDroid dataset is well-suited for the clustering and classification tasks considered in
this research. Specifically, there are~289 dynamic features (i.e., system calls),
200 static features (i.e., permissions, intent filters, metadata),
and we dropped~19 non-numeric features, giving us a total of~470 features.
There are~4 distinct timestamps per data sample~\cite{github};
in our experiments, we use the ``last modified'' timestamp for each sample.

Most of the families in the KronoDroid dataset have only a small number of samples, which makes them
unusable for concept drift experiments. Therefore, we restrict our attention to the following
five malware families, each of which includes a substantial number of samples.
\begin{description}
\item[Airpush/StopSMS] is a Trojan that aggressively pushes ads to the device notification bar~\cite{AirPush}.
For the remainder of this paper, we shorten the name of this family to Airpush.
\item[SMSreg] is ``riskware'' that claims to improve battery life~\cite{SMSReg}.
\item[Malap] is a generic family that includes malicious apps that steal data or install additional malware~\cite{Alejandro}.
\item[Boxer] is a Trojan that sends SMS messages without the user's authorization~\cite{Boxer}.
\item[Agent] is a family of programs that download and install malware~\cite{Agent}.
\end{description}

The number of samples in each family, along with
other relevant details, are provided in Table~\ref{tab:malware_families_intervals}.
In all cases, we use a batch size of~50 in our experiments. 
This approach ensures that we have a sufficient number of 
samples for training and provides a fair comparison 
across different families and models. 

\begin{table}[!htb]
\def\zz{\phantom{0}}
\newsavebox{\mybox}
\sbox{\mybox}{September~13}
\centering
\caption{Malware dataset}\label{tab:malware_families_intervals}
\adjustbox{scale=0.85}{
\begin{tabular}{c|ccll}
\toprule
\multirow{2}{*}{\textbf{Family}} 
	& \multirow{2}{*}{\textbf{Samples}} 
	& \multirow{2}{*}{\textbf{Intervals}}
	& \multicolumn{2}{c}{\textbf{Date range}} \\ \cline{4-5} \\[-2.6ex]
  & & & \multicolumn{1}{c}{\textbf{Begin}} & \multicolumn{1}{c}{\textbf{End}} \\ \midrule
Airpush & \zz7,775 & 154 & February~29, 2008 & \makebox[\wd\mybox][r]{June~23}, 2016 \\ 
SMSreg & \zz5,019 & \zz84 & February~29, 2008 & \makebox[\wd\mybox][r]{November~9}, 2020 \\ 
Malap & \zz4,055 & \zz70 & February~29, 2008 & \makebox[\wd\mybox][r]{November~11}, 2020 \\ 
Boxer & \zz3,597 & \zz69 & February~29, 2008 & \makebox[\wd\mybox][r]{September~13}, 2017 \\ 
Agent & \zz2,934 & \zz46 & February~29, 2008 & \makebox[\wd\mybox][r]{May~6}, 2020 \\ \midrule
Total & 23,380 & 423 & February~29, 2008 & \makebox[\wd\mybox][r]{November~11}, 2020 \\ \bottomrule
\end{tabular}
}
\end{table}

\subsection{Data Preprocessing}

As mentioned above, the KronoDroid dataset includes temporal metadata. 
We segment each family into fixed size batches,
which allows us to study evolution and concept drift within each 
malware family under consideration.

As mentioned, we partition the samples into temporal batches, with~50 samples per batch. 
In addition, we normalize all feature values, which is standard practice when 
training machine learning models.

\subsection{Model Retraining Scenarios}\label{sect:scenarios}


Consider an experiment that we denote as ``Family~$X$ vs Family~$Y$.''
Here, Family~$X$ is the family for which we are trying to detect concept drift, 
while Family~$Y$ simply serves to provide samples for the ``other'' class
in our binary classification experiments. 

The samples of family~$X$ are segmented into temporal batches of~50 samples each.
We denote these batches of~$X$ as~$B_1,B_2,\ldots,B_n$. 
Let~${\cal B}_i = \{B_i,B_{i+1}\}$, for~$i=1,2,\ldots,n-1$.
MiniBatch $K$-Means clustering is applied to each~${\cal B}_i$, and the 
average silhouette coefficient corresponding to~${\cal B}_i$ is denote as~$s_i$. 

The subset of~$B_i$ consisting of its first~25 samples is denoted~$B_i^{t}$ 
while the last~25 samples is denoted~$B_i^{T}$. Randomly select two disjoint
subsets of~25 samples each from Family~$Y$, and denote these samples 
as~$Y_{t}$ and~$Y_{T}$. 
We now describe the three model retraining scenarios.

\subsubsection{Static Training}

Train a model~$M$, which is a binary classifier, to distinguish
between the samples from Family~$X$ in~$B^t_1$ and the Family~$Y$
samples in~$Y_t$.
Test the model~$M$ on each of~$(B^T_i,Y_T)$, for~$i=1,2,\ldots,n$. 
We want to determine the overall accuracy, 
and we will graph the accuracy for each batch. 
If concept drift is occurring in Family~$X$,
we expect the accuracy to decrease when drift occurs.

\subsubsection{Periodic Retraining}

For~$i=1,2,\ldots,n$, train model~$M_i$  
on the labeled samples~$(B^t_i,Y_i)$. 
Then test model~$M_i$ on~$(B^T_i,Y_T)$, for~$i=1,2,\ldots,n$. 
Note that model~$M_1$ in this scenario is the same as model~$M$ in the
static scenario.

As in the static scenario,
we want to determine the overall accuracy, and we will graph the accuracy for each batch.
We expect the accuracy to be higher than the static training scenario,
and the accuracy is likely to be reasonably consistent over the batches.

\subsubsection{Drift-Aware Retraining}
 
In this scenario, a model is retrained only when 
concept drift is detected, and this model is then used to score samples until drift is 
once again detected.
For example, suppose that concept drift is first detected at~${\cal B}_i$ and 
the next drift detection occurs at~${\cal B}_j$, and drift is detected for the final time
at~${\cal B}_k$, where
$$
  1 < i < j < k < n
$$ 
Then model~$M'_1$ is trained on~$(B^t_1,Y_t)$,
model~$M'_2$ is trained on~$(B^t_{i+1},Y_t)$,
model~$M'_3$ is trained on~$(B^t_{j+1},Y_t)$, and
model~$M'_4$ is trained on~$(B^t_{k+1},Y_t)$.
Finally, model~$M'_1$ is used to classify test samples
$$
    (B^T_1,Y_T),\ (B^T_2,Y_T),\ \ldots,\ (B^T_i,Y_T)
$$
model~$M'_2$ is used to classify test samples
$$
    (B^T_{i+1},Y_T),\ (B^T_{i+2},Y_T),\ \ldots,\ (B^T_j,Y_T)
$$
model~$M'_3$ is used to classify test samples
$$
    (B^T_{j+1},Y_T),\ (B^T_{j+2},Y_T),\ \ldots,\ (B^T_k,Y_T)
$$
and model~$M'_4$ is used to classify test samples
$$
    (B^T_{k+1},Y_T),\ (B^T_{k+2},Y_T),\ \ldots,\ (B^T_n,Y_T)
$$
Note that~$M'_1$ in this scenario is the same as~$M_1$ in the periodic retraining scenario,
$M'_2$ is the same as~$M_{i+1}$ in the periodic retraining scenario,
$M'_3$ is the same as~$M_{j+1}$ in the periodic retraining scenario, and
$M'_4$ is the same as~$M_{k+1}$ in the periodic retraining scenario.

As in the previous scenarios, 
we want to determine the overall accuracy, and we will graph the accuracy for each batch.
If we are accurately detecting concept drift, we expect that the results for
this scenario will be similar to that for the periodic retraining scenario.
This approach will also generally be more efficient than the periodic
retraining scenario, since we will train fewer models.

\subsection{Development Environment}

For our development environment we use libraries from Python to 
implement and evaluate our models. Table~\ref{tab:development_environment} 
lists these libraries and how they are used in our experiments.

\begin{table}[!htb]
\centering
\caption{Development environment and libraries}
\label{tab:development_environment}
\adjustbox{scale=0.85}{
\begin{tabular}{c|c}
\toprule
\textbf{Library/Tool} & \textbf{Purpose} \\ \midrule
\multirow{4}{*}{Scikit-learn} & MiniBatch $K$-Means clustering  \\
& Models (MLP, SVM, Random Forest,) \\
& Feature selection (RFE), evaluation metrics \\
& Silhouette score calculation \\ \midrule
\multirow{3}{*}{NumPy} & Computation on large arrays and matrices \\
& Handling malware feature vectors\\
& Calculate distance between samples \\ \midrule
\multirow{2}{*}{Matplotlib} & Visualizations of clustering results \\
& Graph silhouette coefficients \\ \midrule
\multirow{2}{*}{Pandas} & Data manipulation using DataFrames \\
& Preprocessing (filtering, aggregation, missing values) \\ \midrule
\multirow{2}{*}{SciPy (spatial module)} & Compute pairwise distances \\
& $K$-Means and silhouette calculations \\ \midrule
XGBoost & XGBoost classifier \\ \bottomrule
\end{tabular}
}
\end{table}

\subsection{Workflow}

The workflow is divided into four stages. 
The first stage is data collection and preprocessing, which involves downloading the KronoDroid dataset, 
normalizing the features, generating temporal batches for each family.
The second stage is clustering and analysis, where for each malware family, 
we apply MiniBatch $K$-Means to consecutive batches and calculate silhouette coefficients.
In the third stage, we analyze the silhouette coefficients for concept drift detection,
which will be used in the drift-aware retraining scenario. Finally, at model training and evaluation stage, 
we train supervised classification models (MLP, SVM, Random Forest, and XGBoost) 
under each of the three retraining scenarios---static training, periodic retraining, and drift-aware retraining. 
We evaluate the performance of each model under each retraining scenario, where
accuracy is our criteria.

\section{Experiments and Results}\label{chap:ER}

In this section, we present our experimental results. 
We compare how well each retraining scenario performs,
in terms of classification accuracy and discuss the significance of 
our results, with the emphasis on concept drift. 

\subsection{Experimental Setup}\label{sect:setup}

Our four classification models (MLP, SVM, Random Forest, and XGBoost)
are evaluated under each of the three retraining scenarios discussed in
Section~\ref{sect:scenarios}. Recall that these three scenarios are denoted
as static training, periodic retraining, and drift-aware retraining.

For the drift-aware retraining scenario, we need to set a threshold that specifies 
when concept drift has been detected. Let~$d_i = |s_i - s_{i-1}|$,
where~$s_i$ is the average silhouette coefficient for the
clustering of the~$i^{\thth}$ temporal subset.\footnote{Recall that
in the notation of Section~\ref{sect:scenarios}, we split the 
family into temporal batches~$B_i$ of~50 samples each,
and then cluster two consecutive of these batches, which we
denote as~${\cal B}_i = \{B_i,B_{i+1}\}$.}
Based on preliminary experiments,
we use a threshold of~$d_i > 0.05$,
to trigger retraining in the drift-aware scenario.

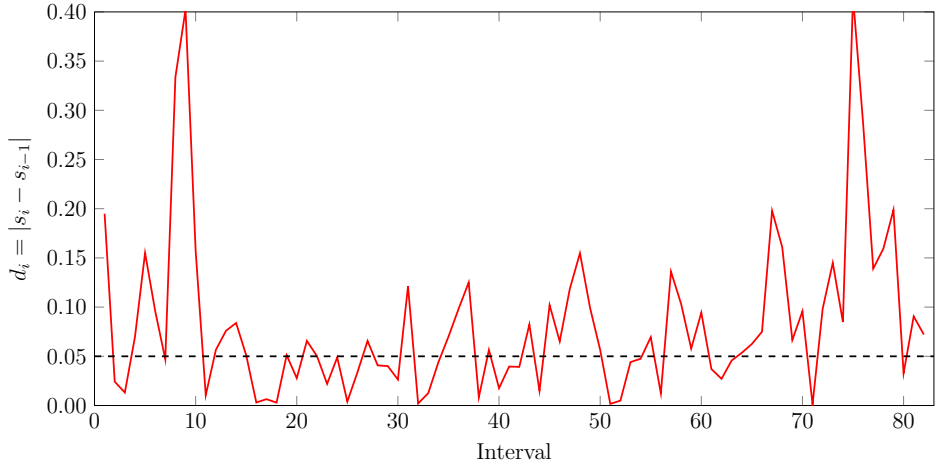
\begin{figure}[!htb]
    \centering
    \begin{tikzpicture}[scale=0.85]
\begin{axis}[xmin=0, xmax=83, ymin=0.0, ymax=0.40,
		   width=0.95\textwidth,
		   height=0.50\textwidth,
	 	   x tick label style={scale=0.85,
   		 	/pgf/number format/.cd,
			/pgf/number format/1000 sep={},
   			fixed,
   			fixed zerofill,
    			precision=0
		   },
		   x label style={scale=0.85},
	 	   y tick label style={scale=0.85,
    		 	/pgf/number format/.cd,
   			fixed,
   			fixed zerofill,
    			precision=2
		    },
		   y label style={scale=0.85},
                    xtick={0,10,20,30,40,50,60,70,80},
                    ytick={0.0,0.05,0.1,0.15,0.2,0.25,0.3,0.35,0.40},
                    xlabel={Interval},
                    ylabel={$d_i = |s_i-s_{i-1}|$}] 
\addplot[color=red, thick, mark=none] coordinates { 
(1, 0.19492417598194195)
(2, 0.02421630403404129)
(3, 0.013159597445400656)
(4, 0.06965175701982053)
(5, 0.15500813019041423)
(6, 0.09606754581647081)
(7, 0.04704998583948056)
(8, 0.33382045304677926)
(9, 0.40225711640420164)
(10, 0.1574214692555908)
(11, 0.010426351849826265)
(12, 0.05653588736054699)
(13, 0.07593915375586091)
(14, 0.08390063190651581)
(15, 0.05103825378719115)
(16, 0.0030474564412527327)
(17, 0.006468173601820171)
(18, 0.0029266682414371137)
(19, 0.051147301533362693)
(20, 0.027761515412150606)
(21, 0.06570812653169728)
(22, 0.05019460492776051)
(23, 0.021958902215157894)
(24, 0.04885845534022995)
(25, 0.004049175211414541)
(26, 0.033739654431537314)
(27, 0.0656539311375735)
(28, 0.04078250557139049)
(29, 0.04007012068181684)
(30, 0.026370733162633164)
(31, 0.12130659565402827)
(32, 0.0020292606203016206)
(33, 0.012579578597787633)
(34, 0.04384784611914527)
(35, 0.07001242994430396)
(36, 0.09819183788940966)
(37, 0.12508541404490003)
(38, 0.008305741848221665)
(39, 0.05652461998276401)
(40, 0.017557345931993595)
(41, 0.039597682766430886)
(42, 0.03924732285180357)
(43, 0.08232914493032531)
(44, 0.014976890822478384)
(45, 0.10206200791363729)
(46, 0.06510981728448723)
(47, 0.1186864240563743)
(48, 0.15479967364009684)
(49, 0.09921438780981828)
(50, 0.05645796744365028)
(51, 0.0016607128323418707)
(52, 0.004951929498276664)
(53, 0.04415484354920629)
(54, 0.0474903382455501)
(55, 0.06947710539943586)
(56, 0.012349642628403101)
(57, 0.13607285590844798)
(58, 0.10334151994520602)
(59, 0.05785866333159434)
(60, 0.09440759847143476)
(61, 0.03698299086124884)
(62, 0.02712922098084186)
(63, 0.04573642872114542)
(64, 0.05353863188368585)
(65, 0.06251874340862895)
(66, 0.07498313157798442)
(67, 0.19790779427204475)
(68, 0.16090091314584543)
(69, 0.06659295369503149)
(70, 0.0957894571335775)
(71, 0.0010303389979941513)
(72, 0.09841065154690659)
(73, 0.1449933448067265)
(74, 0.08487565009416975)
(75, 0.41548383359571484)
(76, 0.2865690712823414)
(77, 0.13913088368847754)
(78, 0.1593400204112415)
(79, 0.19876442805523287)
(80, 0.03281783504150004)
(81, 0.0905762476462913)
(82, 0.07215184321250628)
};
\addplot[color=black, thick, dashed] table[row sep = crcr]{0 0.05 \\ 152 0.05 \\};
\end{axis}
\end{tikzpicture}
    \caption{Silhouette score differences for SMSreg}\label{fig:SMS_sil}
\end{figure}

Analysis of silhouette coefficients indicate that some families
exhibit relatively stable results, and other families yield more dynamic results. 
For example, the values of~$d_i$ for SMSreg are given in the form of
a line graph in Figure~\ref{fig:SMS_sil}. In this case, 
we observe that~$d_i$ ranges from essentially~0 to~0.4,
with relatively stable values from interval~10 to interval~73.


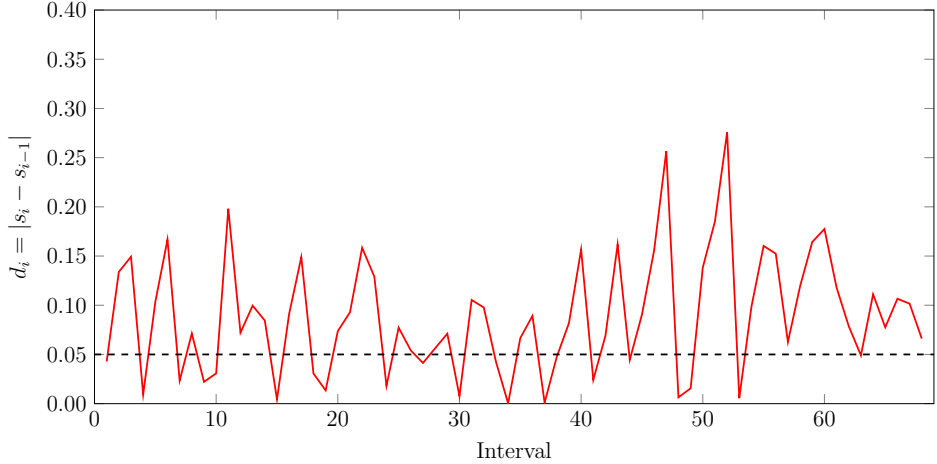
\begin{figure}[!htb]
    \centering
    \begin{tikzpicture}[scale=0.85]
\begin{axis}[xmin=0, xmax=69, ymin=0.0, ymax=0.40,
		   width=0.95\textwidth,
		   height=0.50\textwidth,
	 	   x tick label style={scale=0.85,
   		 	/pgf/number format/.cd,
			/pgf/number format/1000 sep={},
   			fixed,
   			fixed zerofill,
    			precision=0
		   },
		   x label style={scale=0.85},
	 	   y tick label style={scale=0.85,
    		 	/pgf/number format/.cd,
   			fixed,
   			fixed zerofill,
    			precision=2
		    },
		   y label style={scale=0.85},
                    xtick={0,10,20,30,40,50,60,70},
                    ytick={0.0,0.05,0.1,0.15,0.2,0.25,0.3,0.35,0.40},
                    xlabel={Interval},
                    ylabel={$d_i = |s_i-s_{i-1}|$}] 
\addplot[color=red, thick, mark=none] coordinates { 
(1, 0.0428647063197134)
(2, 0.133857457692662)
(3, 0.14928870893279003)
(4, 0.00936746806274319)
(5, 0.10342643633690793)
(6, 0.16675259677434062)
(7, 0.023401369406084366)
(8, 0.07131265368379289)
(9, 0.022206570771396927)
(10, 0.030750477192786008)
(11, 0.19807146985240776)
(12, 0.0721468233726778)
(13, 0.09960025226796859)
(14, 0.08445222191282115)
(15, 0.0042874398923413715)
(16, 0.0912729105692594)
(17, 0.1488724288892252)
(18, 0.03060624554185104)
(19, 0.013471407994512075)
(20, 0.07372206448521645)
(21, 0.09285789329264704)
(22, 0.1583532632880425)
(23, 0.12916942472470389)
(24, 0.018071338012850047)
(25, 0.07722876335628542)
(26, 0.054068655315311465)
(27, 0.04133604807895602)
(28, 0.05624483784184292)
(29, 0.07108840247692416)
(30, 0.007693784248144486)
(31, 0.10537419793037528)
(32, 0.0977133954254375)
(33, 0.04258576223607341)
(34, 0.00025040966979061885)
(35, 0.06659432951700645)
(36, 0.08921871255649805)
(37, 0.0009932529878252105)
(38, 0.04760863158980494)
(39, 0.08189666536292534)
(40, 0.1566317120890821)
(41, 0.024161254629494866)
(42, 0.06902718221561716)
(43, 0.16175268332064108)
(44, 0.04450389922912257)
(45, 0.09052857880973292)
(46, 0.15568551039475514)
(47, 0.25651628649768615)
(48, 0.006449829555552955)
(49, 0.015519242596729987)
(50, 0.13830247934608336)
(51, 0.185218415721767)
(52, 0.2757395846292055)
(53, 0.005569466512181209)
(54, 0.09853661136652192)
(55, 0.16030350942327076)
(56, 0.15239088081270186)
(57, 0.06262397994633184)
(58, 0.11940108309289876)
(59, 0.16430629004399872)
(60, 0.17747860865794918)
(61, 0.11782981641120105)
(62, 0.07869208179315756)
(63, 0.049048227152612034)
(64, 0.111128496320681)
(65, 0.07755529493916052)
(66, 0.10659116803132201)
(67, 0.10161245422717535)
(68, 0.06610930498832596)
};
\addplot[color=black, thick, dashed] table[row sep = crcr]{0 0.05 \\ 152 0.05 \\};
\end{axis}
\end{tikzpicture}
    \caption{Silhouette score differences for Malap}\label{fig:malap_sil}
\end{figure}

As another example, the values of~$d_i$ for Malap are given in Figure~\ref{fig:malap_sil}. 
For this family, we observe that the absolute differences between consecutive
pairs of average silhouette coefficients (i.e., $d_i$) vary over a smaller range than
for SMSreg, but with a more consistent variation throughout all intervals.


Analogous graphs to those in Figures~\ref{fig:SMS_sil} and~\ref{fig:malap_sil}
for the Agent, Airpush, and Boxer families are
given in the Appendix in Figures~\ref{fig:agent_sil}, 
\ref{fig:airpush_sil}, and~\ref{fig:boxer_sil}, respectively. 
We note that among the five families considered,
Malap exceeds the threshold of~$d_i > 0.05$
with the highest relative frequency, while
Boxer exceeds the threshold with the lowest
relative frequency. We provide more details
in Section~\ref{sect:disc}, below.

\subsection{Hyperparameter Tuning}

Table~\ref{tab:parameters} lists the hyperparameters tested for each model.
The hyperparameters selected for MinBatch $K$-Means are highlighted in boldface. 
For the classification models, a separate grid search was performed over the listed 
hyperparameters for each family.

\begin{table}[!htb]
\centering
\caption{Hyperparameters for clustering and classification models}
\label{tab:parameters}
\adjustbox{scale=0.85}{
\begin{tabular}{c|cc}
\toprule
\textbf{Model} & \textbf{Hyperparameters} & \textbf{Tested values} \\ \midrule
\multirow{3}{*}{MiniBatch $K$-Means} & Batch sizes & \textbf{50}, 100, 150 \\ 
 & MiniBatch sizes & 15, \textbf{20}, 25  \\ 
 & Drift detection threshold & 0.01, 0.03, \textbf{0.05} 0.10 \\ \midrule
\multirow{4}{*}{MLP} & \texttt{hidden\_layer\_sizes} & (50,), (100,), (100,75,75) \\ 
 & \texttt{max\_iter} & 200, 300, 500 \\ 
 & \texttt{activation} & relu, tanh \\ 
 & \texttt{solver} & adam, sgd \\ \midrule
\multirow{4}{*}{SVM} & \texttt{kernel} & linear \\ 
& \texttt{penalty} & L1, L2 \\ 
 & \texttt{C} & 0.01, 0.1, 1.0 \\ 
 & \texttt{tol} & 0.001, 0.0001 \\ \midrule
\multirow{4}{*}{Random Forest} & \texttt{n\_estimators} & 10, 50, 200 \\ 
 & \texttt{max\_depth} & 10, 20, 30 \\ 
 & \texttt{min\_samples\_split} & 2, 5, 10 \\ 
 & \texttt{min\_samples\_leaf} & 1, 5, 10 \\ \midrule
\multirow{3}{*}{XGBoost} & \texttt{n\_estimators} & 50, 100, 200 \\ 
 & \texttt{max\_depth} & 3, 4, 5 \\ 
 & \texttt{learning\_rate} & 0.01, 0.1, 0.2 \\ \bottomrule
\end{tabular}
}
\end{table}

\subsection{Retraining Scenario Results}

This section contains detailed results on the performance of each of the four learning models 
under each of the three retraining scenarios. Note that every learning model is trained and tested
on a binary classification
problem, where ``Family~$X$ vs Family~$Y$'' indicates that Family~$X$ is segmented
into temporal subsets, with models trained, retrained, and tested 
based on these temporal subsets, with
Family~$Y$ simply serving as (fixed) ``not Family~$X$'' data for each trained model.
For each of our five malware families serves as Family~$X$ 
with selected choices for Family~$Y$, 
as some families are inherently easy to distinguish from each other, 
and the easiest cases would tend to mask the effect of concept drift.

\subsubsection{Static Training Results}

\begin{figure}[!htb]
    \centering
    \begin{tikzpicture}[scale=0.8, every node/.style={scale=1.0}]
\pgfkeys{/pgf/number format/.cd,1000 sep={}}
\begin{axis}[
        width  = 1.1*\textwidth,
        height = 8.25cm,
        ymin=0.0,ymax=1.02,
        ytick={0.0, 0.2, 0.4, 0.6, 0.8, 1.0},
        major x tick style = transparent,
        ybar=5*\pgflinewidth,
        bar width=2.25pt,
        ylabel = {Accuracy},
        ylabel style = {scale = 0.9},
        symbolic x coords={
        Airpush vs SMSreg, 
        Airpush vs Malap, 
        Airpush vs Boxer, 
        Airpush vs Agent, 
        SMSreg vs Airpush, 
        SMSreg vs Malap, 
        SMSreg vs Boxer, 
        SMSreg vs Agent, 
        Malap vs Airpush, 
        Malap vs SMSreg, 
        Malap vs Boxer, 
        Malap vs Agent, 
        Boxer vs Airpush, 
        Boxer vs SMSReg, 
        Boxer vs Malap, 
        Boxer vs Agent, 
        Agent vs Airpush, 
        Agent vs SMSreg, 
        Agent vs Malap, 
        Agent vs Boxer
        },
        xticklabels={
        Airpush vs SMSreg, 
        Airpush vs Malap, 
        Airpush vs Boxer, 
        Airpush vs Agent, 
        SMSreg vs Airpush, 
        SMSreg vs Malap, 
        SMSreg vs Boxer, 
        SMSreg vs Agent, 
        Malap vs Airpush, 
        Malap vs SMSreg, 
        Malap vs Boxer, 
        Malap vs Agent, 
        Boxer vs Airpush, 
        Boxer vs SMSReg, 
        Boxer vs Malap, 
        Boxer vs Agent, 
        Agent vs Airpush, 
        Agent vs SMSreg, 
        Agent vs Malap, 
        Agent vs Boxer
        },
	y tick label style={scale=0.9,
    		/pgf/number format/.cd,
   		fixed,
   		fixed zerofill,
    		precision=2},
        xtick = data,
        x tick label style={scale=0.8,
        		rotate=60,
		anchor=north east,
		inner sep=0mm
		},
        enlarge x limits=0.0325,
        legend cell align=left,
        legend pos=south east,
        legend style={nodes={scale=0.85},
        },
]
\addplot [fill=blue,opacity=1.00]
coordinates {
(Airpush vs SMSreg, 0.5768)
(Airpush vs Malap, 0.4968)
(Airpush vs Boxer, 0.6781)
(Airpush vs Agent, 0.5195)
(SMSreg vs Airpush, 0.5462)
(SMSreg vs Malap, 0.4612)
(SMSreg vs Boxer, 0.6055)
(SMSreg vs Agent, 0.4590)
(Malap vs Airpush, 0.5046)
(Malap vs SMSreg, 0.5617)
(Malap vs Boxer, 0.6669)
(Malap vs Agent, 0.4211)
(Boxer vs Airpush, 0.5806)
(Boxer vs SMSReg, 0.5214)
(Boxer vs Malap, 0.3870)
(Boxer vs Agent, 0.5214)
(Agent vs Airpush, 0.4878)
(Agent vs SMSreg, 0.4717)
(Agent vs Malap, 0.4574)
(Agent vs Boxer, 0.7226)
};
\addlegendentry{MLP}
\addplot [fill=yellow,opacity=1.00]
coordinates {
(Airpush vs SMSreg, 0.6731)
(Airpush vs Malap, 0.7790)
(Airpush vs Boxer, 0.9436)
(Airpush vs Agent, 0.7942)
(SMSreg vs Airpush, 0.9059)
(SMSreg vs Malap, 0.8098)
(SMSreg vs Boxer, 0.9855)
(SMSreg vs Agent, 0.5998)
(Malap vs Airpush, 0.5871)
(Malap vs SMSreg, 0.4883)
(Malap vs Boxer, 0.8780)
(Malap vs Agent, 0.4657)
(Boxer vs Airpush, 0.8632)
(Boxer vs SMSReg, 0.8386)
(Boxer vs Malap, 0.7765)
(Boxer vs Agent, 0.7849)
(Agent vs Airpush, 0.8265)
(Agent vs SMSreg, 0.6009)
(Agent vs Malap, 0.6752)
(Agent vs Boxer, 0.8230)
};
\addlegendentry{SVM}
\addplot [fill=green,opacity=1.00]
coordinates {
(Airpush vs SMSreg, 0.8452)
(Airpush vs Malap, 0.9196)
(Airpush vs Boxer, 0.9405)
(Airpush vs Agent, 0.8626)
(SMSreg vs Airpush, 0.8317)
(SMSreg vs Malap, 0.5460)
(SMSreg vs Boxer, 0.9969)
(SMSreg vs Agent, 0.6643)
(Malap vs Airpush, 0.6417)
(Malap vs SMSreg, 0.4800)
(Malap vs Boxer, 0.9866)
(Malap vs Agent, 0.5729)
(Boxer vs Airpush, 0.9440)
(Boxer vs SMSReg, 0.9119)
(Boxer vs Malap, 0.8812)
(Boxer vs Agent, 0.9559)
(Agent vs Airpush, 0.8922)
(Agent vs SMSreg, 0.4974)
(Agent vs Malap, 0.8087)
(Agent vs Boxer, 0.9748)
};
\addlegendentry{Random Forest}
\addplot [fill=red,opacity=1.00]
coordinates {
(Airpush vs SMSreg, 0.4842)
(Airpush vs Malap, 0.9252)
(Airpush vs Boxer, 0.9209)
(Airpush vs Agent, 0.8742)
(SMSreg vs Airpush, 0.5288)
(SMSreg vs Malap, 0.5069)
(SMSreg vs Boxer, 0.9857)
(SMSreg vs Agent, 0.4981)
(Malap vs Airpush, 0.8823)
(Malap vs SMSreg, 0.6831)
(Malap vs Boxer, 0.8997)
(Malap vs Agent, 0.8223)
(Boxer vs Airpush, 0.9562)
(Boxer vs SMSReg, 0.9070)
(Boxer vs Malap, 0.6180)
(Boxer vs Agent, 0.9342)
(Agent vs Airpush, 0.9300)
(Agent vs SMSreg, 0.6117)
(Agent vs Malap, 0.6504)
(Agent vs Boxer, 0.9183)
};
\addlegendentry{XGBoost}
\end{axis}
\end{tikzpicture}
    \caption{Accuracies across all models for static training scenario}\label{fig:caseA_All}
\end{figure}
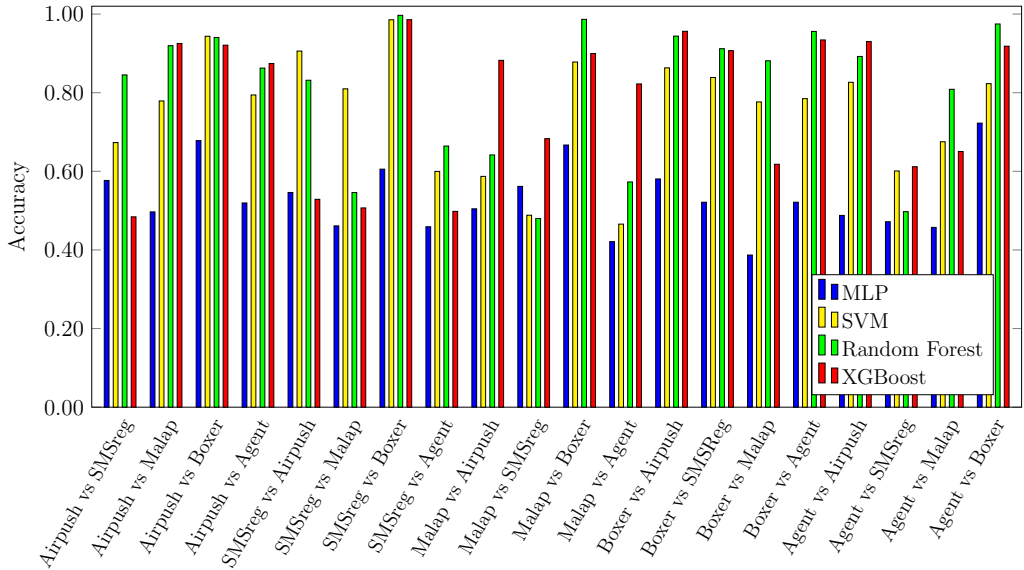

Recall that in the static training scenario, models are trained on the initial temporal subset,
and this model is then used to classify all remaining test samples.
This scenario represents a situation where concept drift is ignored, and it
serves as a baseline. The results of all of our static training experiments are 
summarized in the form of a bar graph in Figure~\ref{fig:caseA_All}. 

From Figure~\ref{fig:caseA_All},
we observe that Random Forest and XGBoost 
generally perform best, with MLP clearly the worst.
The poor results for MLP are likely due to a lack of training data,
since we only have~50 samples per batch.
We note that some families are much easier to distinguish
than others; for example the results for SMSreg vs Boxer are
among the best, while SMSreg vs Agent is among the worst-performing.

\subsubsection{Periodic Retraining Results}

\begin{figure}[!htb]
    \centering
    \begin{tikzpicture}[scale=0.8, every node/.style={scale=1.0}]
\pgfkeys{/pgf/number format/.cd,1000 sep={}}
\begin{axis}[
        width  = 1.1*\textwidth,
        height = 8.25cm,
        ymin=0.0,ymax=1.02,
        ytick={0.0, 0.2, 0.4, 0.6, 0.8, 1.0},
        major x tick style = transparent,
        ybar=5*\pgflinewidth,
        bar width=2.25pt,
        ylabel = {Accuracy},
        ylabel style = {scale = 0.9},
        symbolic x coords={
        Airpush vs SMSreg, 
        Airpush vs Malap, 
        Airpush vs Boxer, 
        Airpush vs Agent, 
        SMSreg vs Airpush, 
        SMSreg vs Malap, 
        SMSreg vs Boxer, 
        SMSreg vs Agent, 
        Malap vs Airpush, 
        Malap vs SMSreg, 
        Malap vs Boxer, 
        Malap vs Agent, 
        Boxer vs Airpush, 
        Boxer vs SMSReg, 
        Boxer vs Malap, 
        Boxer vs Agent, 
        Agent vs Airpush, 
        Agent vs SMSreg, 
        Agent vs Malap, 
        Agent vs Boxer
        },
        xticklabels={
        Airpush vs SMSreg, 
        Airpush vs Malap, 
        Airpush vs Boxer, 
        Airpush vs Agent, 
        SMSreg vs Airpush, 
        SMSreg vs Malap, 
        SMSreg vs Boxer, 
        SMSreg vs Agent, 
        Malap vs Airpush, 
        Malap vs SMSreg, 
        Malap vs Boxer, 
        Malap vs Agent, 
        Boxer vs Airpush, 
        Boxer vs SMSReg, 
        Boxer vs Malap, 
        Boxer vs Agent, 
        Agent vs Airpush, 
        Agent vs SMSreg, 
        Agent vs Malap, 
        Agent vs Boxer
        },
	y tick label style={scale=0.9,
    		/pgf/number format/.cd,
   		fixed,
   		fixed zerofill,
    		precision=2},
        xtick = data,
        x tick label style={scale=0.8,
        		rotate=60,
		anchor=north east,
		inner sep=0mm
		},
        enlarge x limits=0.0325,
        legend cell align=left,
        legend pos=south east,
        legend style={nodes={scale=0.85},
        },
]
\addplot [fill=blue,opacity=1.00]
coordinates {
(Airpush vs SMSreg, 0.6038)
(Airpush vs Malap,  0.5138)
(Airpush vs Boxer, 0.7039)
(Airpush vs Agent, 0.5386)
(SMSreg vs Airpush, 0.7912)
(SMSreg vs Malap, 0.6636)
(SMSreg vs Boxer, 0.8443)
(SMSreg vs Agent, 0.7236)
(Malap vs Airpush, 0.5946)
(Malap vs SMSreg, 0.6129)
(Malap vs Boxer, 0.7046)
(Malap vs Agent, 0.5460)
(Boxer vs Airpush, 0.7713)
(Boxer vs SMSReg, 0.7058)
(Boxer vs Malap, 0.6643)
(Boxer vs Agent, 0.7501)
(Agent vs Airpush, 0.6378)
(Agent vs SMSreg, 0.6083)
(Agent vs Malap, 0.5426)
(Agent vs Boxer, 0.7539)
};
\addlegendentry{MLP}
\addplot [fill=yellow,opacity=1.00]
coordinates {
(Airpush vs SMSreg, 0.6899)
(Airpush vs Malap, 0.8196)
(Airpush vs Boxer, 0.9664)
(Airpush vs Agent, 0.8042)
(SMSreg vs Airpush, 0.9383)
(SMSreg vs Malap, 0.8848)
(SMSreg vs Boxer, 0.9821)
(SMSreg vs Agent, 0.8562)
(Malap  vs  Airpush,  0.8983)
(Malap vs SMSreg, 0.7791)
(Malap vs Boxer, 0.9463)
(Malap vs Agent, 0.8266)
(Boxer vs Airpush, 0.9483)
(Boxer vs SMSReg, 0.8991)
(Boxer vs Malap, 0.8209)
(Boxer vs Agent, 0.9125)
(Agent vs Airpush, 0.8222)
(Agent vs SMSreg, 0.6357)
(Agent vs Malap, 0.6887)
(Agent vs Boxer, 0.9200)
};
\addlegendentry{SVM}
\addplot [fill=green,opacity=1.00]
coordinates {
(Airpush vs SMSreg, 0.7891)
(Airpush vs Malap, 0.9442)
(Airpush vs Boxer,  0.9813)
(Airpush vs Agent, 0.9087)
(SMSreg vs Airpush, 0.9788)
(SMSreg vs Malap, 0.9531)
(SMSreg vs Boxer, 0.9907)
(SMSreg vs Agent, 0.9426)
(Malap vs Airpush, 0.9531)
(Malap vs SMSreg, 0.8774)
(Malap vs Boxer, 0.9711)
(Malap vs Agent, 0.8909)
(Boxer vs Airpush, 0.9693)
(Boxer vs SMSReg, 0.9507)
(Boxer vs Malap, 0.9235)
(Boxer vs Agent, 0.9629)
(Agent vs Airpush, 0.8883)
(Agent vs SMSreg, 0.6517)
(Agent vs Malap, 0.8209)
(Agent vs Boxer, 0.9696)
};
\addlegendentry{Random Forest}
\addplot [fill=red,opacity=1.00]
coordinates {
(Airpush vs SMSreg, 0.7471)
(Airpush vs Malap, 0.9442)
(Airpush vs Boxer, 0.9705)
(Airpush vs Agent, 0.8832)
(SMSreg vs Airpush, 0.9638)
(SMSreg vs Malap, 0.9345)
(SMSreg vs Boxer, 0.9733)
(SMSreg vs Agent, 0.9412)
(Malap vs Airpush, 0.9260)
(Malap vs SMSreg, 0.8197)
(Malap vs Boxer, 0.9509)
(Malap vs Agent, 0.8729)
(Boxer vs Airpush, 0.9617)
(Boxer vs SMSReg, 0.9258)
(Boxer vs Malap, 0.8464)
(Boxer vs Agent, 0.9249)
(Agent vs Airpush, 0.8787)
(Agent vs SMSreg, 0.6843)
(Agent vs Malap, 0.7274)
(Agent vs Boxer,  0.9374)
};
\addlegendentry{XGBoost}
\end{axis}
\end{tikzpicture}
    \caption{Accuracies across all models for periodic retraining scenario}\label{fig:caseB_All}
\end{figure}
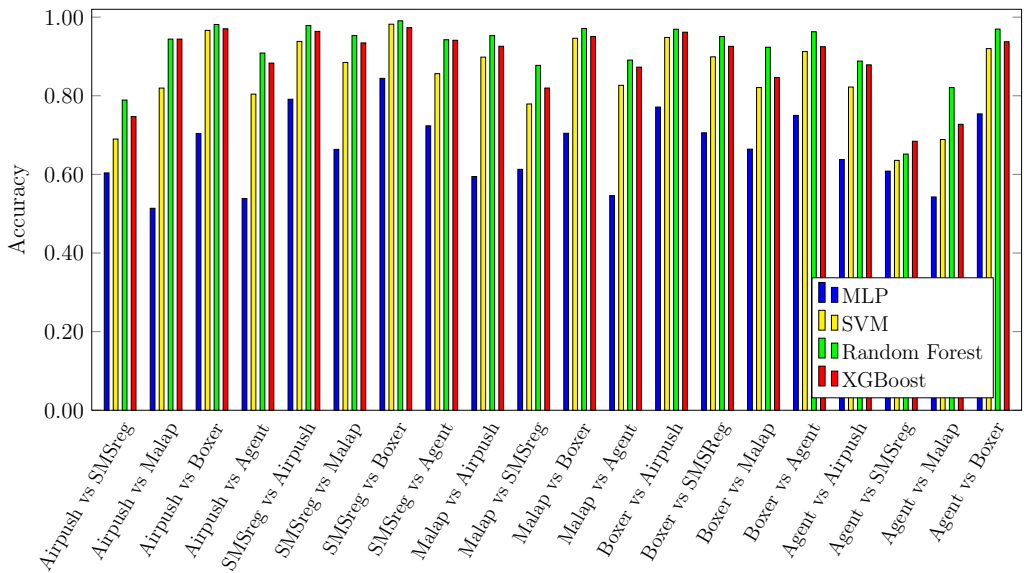

In periodic retraining, models are updated at each interval, 
regardless of whether drift is detected. The results for these experiments
are given in Figure~\ref{fig:caseB_All}.

Comparing Figures~\ref{fig:caseA_All} and~\ref{fig:caseB_All}, 
we observe that the results for the periodic retraining scenario
are, on average, much better than those for the static training scenario.
This is not surprising, as any cases where significant concept drift occurs
will benefit from periodic retraining.

\subsubsection{Drift-Aware Retraining Results}

\begin{figure}[!htb]
    \centering
    \begin{tikzpicture}[scale=0.8, every node/.style={scale=1.0}]
\pgfkeys{/pgf/number format/.cd,1000 sep={}}
\begin{axis}[
        width  = 1.1*\textwidth,
        height = 8.25cm,
        ymin=0.0,ymax=1.02,
        ytick={0.0, 0.2, 0.4, 0.6, 0.8, 1.0},
        major x tick style = transparent,
        ybar=5*\pgflinewidth,
        bar width=2.25pt,
        ylabel = {Accuracy},
        ylabel style = {scale = 0.9},
        symbolic x coords={
        Airpush vs SMSreg, 
        Airpush vs Malap, 
        Airpush vs Boxer, 
        Airpush vs Agent, 
        SMSreg vs Airpush, 
        SMSreg vs Malap, 
        SMSreg vs Boxer, 
        SMSreg vs Agent, 
        Malap vs Airpush, 
        Malap vs SMSreg, 
        Malap vs Boxer, 
        Malap vs Agent, 
        Boxer vs Airpush, 
        Boxer vs SMSReg, 
        Boxer vs Malap, 
        Boxer vs Agent, 
        Agent vs Airpush, 
        Agent vs SMSreg, 
        Agent vs Malap, 
        Agent vs Boxer
        },
        xticklabels={
        Airpush vs SMSreg, 
        Airpush vs Malap, 
        Airpush vs Boxer, 
        Airpush vs Agent, 
        SMSreg vs Airpush, 
        SMSreg vs Malap, 
        SMSreg vs Boxer, 
        SMSreg vs Agent, 
        Malap vs Airpush, 
        Malap vs SMSreg, 
        Malap vs Boxer, 
        Malap vs Agent, 
        Boxer vs Airpush, 
        Boxer vs SMSReg, 
        Boxer vs Malap, 
        Boxer vs Agent, 
        Agent vs Airpush, 
        Agent vs SMSreg, 
        Agent vs Malap, 
        Agent vs Boxer
        },
	y tick label style={scale=0.9,
    		/pgf/number format/.cd,
   		fixed,
   		fixed zerofill,
    		precision=2},
        xtick = data,
        x tick label style={scale=0.8,
        		rotate=60,
		anchor=north east,
		inner sep=0mm
		},
        enlarge x limits=0.0325,
        legend cell align=left,
        legend pos=south east,
        legend style={nodes={scale=0.85},
        },
]
\addplot [fill=blue,opacity=1.00]
coordinates {
(Airpush vs SMSreg, 0.5872)
(Airpush vs Malap,  0.4936)
(Airpush vs Boxer, 0.6956)
(Airpush vs Agent, 0.5240)
(SMSreg vs Airpush, 0.7833)
(SMSreg vs Malap, 0.6729)
(SMSreg vs Boxer, 0.8367)
(SMSreg vs Agent, 0.7217)
(Malap vs Airpush, 0.5854)
(Malap vs SMSreg, 0.6111)
(Malap vs Boxer, 0.7017)
(Malap vs Agent, 0.5346)
(Boxer vs Airpush, 0.7472)
(Boxer vs SMSReg, 0.6809)
(Boxer vs Malap, 0.6400)
(Boxer vs Agent, 0.7333)
(Agent vs Airpush, 0.6334)
(Agent vs SMSreg, 0.6022)
(Agent vs Malap, 0.5017)
(Agent vs Boxer, 0.7509)
};
\addlegendentry{MLP}
\addplot [fill=yellow,opacity=1.00]
coordinates {
(Airpush vs SMSreg, 0.6813)
(Airpush vs Malap, 0.8086)
(Airpush vs Boxer, 0.9644)
(Airpush vs Agent, 0.7883)
(SMSreg vs Airpush, 0.9121)
(SMSreg vs Malap, 0.8781)
(SMSreg vs Boxer, 0.9588)
(SMSreg vs Agent, 0.8186)
(Malap vs Airpush, 0.8906)
(Malap vs SMSreg, 0.7774)
(Malap vs Boxer, 0.9511)
(Malap vs Agent, 0.8220)
(Boxer vs Airpush, 0.9464)
(Boxer vs SMSReg, 0.8954)
(Boxer vs Malap, 0.8174)
(Boxer vs Agent, 0.9017)
(Agent vs Airpush, 0.8039)
(Agent vs SMSreg, 0.6165)
(Agent vs Malap, 0.6800)
(Agent vs Boxer, 0.9400)
};
\addlegendentry{SVM}
\addplot [fill=green,opacity=1.00]
coordinates {
(Airpush vs SMSreg, 0.7726)
(Airpush vs Malap, 0.9391)
(Airpush vs Boxer, 0.9788)
(Airpush vs Agent, 0.9009)
(SMSreg vs Airpush, 0.9743)
(SMSreg vs Malap, 0.9436)
(SMSreg vs Boxer, 0.9898)
(SMSreg vs Agent, 0.9367)
(Malap vs Airpush, 0.9549)
(Malap vs SMSreg, 0.8800)
(Malap vs Boxer, 0.9777)
(Malap vs Agent, 0.8951)
(Boxer vs Airpush, 0.9701)
(Boxer vs SMSReg, 0.9499)
(Boxer vs Malap, 0.9220)
(Boxer vs Agent,  0.9699)
(Agent vs Airpush, 0.8843)
(Agent vs SMSreg, 0.6487)
(Agent vs Malap, 0.8109)
(Agent vs Boxer, 0.9817)
};
\addlegendentry{Random Forest}
\addplot [fill=red,opacity=1.00]
coordinates {
(Airpush vs SMSreg, 0.7378)
(Airpush vs Malap, 0.9348)
(Airpush vs Boxer, 0.9632)
(Airpush vs Agent, 0.8764)
(SMSreg vs Airpush, 0.9612)
(SMSreg vs Malap, 0.9307)
(SMSreg vs Boxer, 0.9717)
(SMSreg vs Agent, 0.9364)
(Malap vs Airpush, 0.9297)
(Malap vs SMSreg, 0.8251)
(Malap vs Boxer, 0.9614)
(Malap vs Agent, 0.8740)
(Boxer vs Airpush, 0.9655)
(Boxer vs SMSReg, 0.9145)
(Boxer vs Malap, 0.8464)
(Boxer vs Agent, 0.9151)
(Agent vs Airpush, 0.8783)
(Agent vs SMSreg, 0.6748)
(Agent vs Malap, 0.7217)
(Agent vs Boxer, 0.9252)
};
\addlegendentry{XGBoost}
\end{axis}
\end{tikzpicture}
    \caption{Accuracies across all models for drift-aware scenario}\label{fig:caseC_All}
\end{figure}
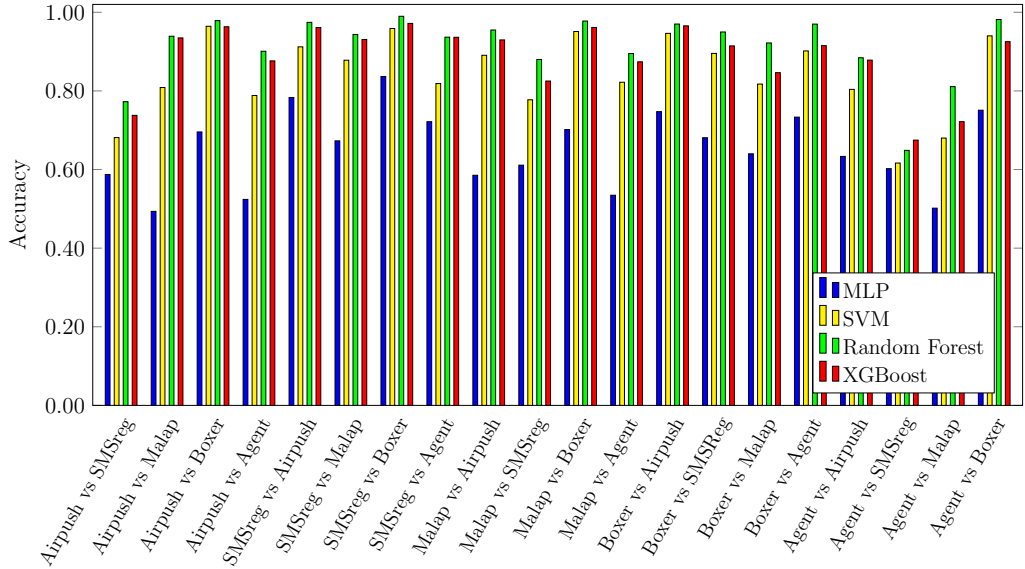

Under the drift-aware retraining scenario, 
models are only updated when concept drift is detected,
as determined by the changes in the average silhouette coefficient
when MiniBatch $K$-Means is applied to consecutive temporal batches,
as discussed in Section~\ref{sect:scenarios}.
The results for this scenario are given in Figure~\ref{fig:caseC_All}.

Comparing Figures~\ref{fig:caseB_All} and~\ref{fig:caseC_All}, we observe
that the results for the drift-aware scenario appear to be comparable to those
obtained in the periodic retraining scenario. This indicates that we 
are detecting concept drift.

\subsection{Discussion}\label{sect:disc}

\begin{figure}[!htb]
    \centering
    \begin{tikzpicture}[scale=0.8, every node/.style={scale=1.0}]
\pgfkeys{/pgf/number format/.cd,1000 sep={}}
\begin{axis}[
        width  = 0.8*\textwidth,
        height = 8.25cm,
        ymin=0.475,ymax=1.02,
        ytick={0.50, 0.60, 0.70, 0.80, 0.90, 1.00},
        major x tick style = transparent,
        ybar=5*\pgflinewidth,
        bar width=15.5pt,
        ylabel = {Accuracy},
        ylabel style = {scale = 0.9},
        symbolic x coords={C, A, B, D},
        xticklabels={MLP,  SVM, Random Forest, XGBoost},
 	y tick label style={scale=0.9,
    		/pgf/number format/.cd,
   		fixed,
   		fixed zerofill,
    		precision=2},
        xtick = data,
        x tick label style={scale=0.8,
		},
        nodes near coords,
        every node near coord/.append style={rotate=90, scale=0.8,
        								   anchor=west, 
								   /pgf/number format/.cd,
								   fixed,
								   fixed zerofill,
								   precision=4},
        enlarge x limits=0.175,
        legend cell align=left,
        legend pos=south east,
        legend style={nodes={scale=0.85},
                column sep=1ex
        },
]
\addplot [fill=green,opacity=1.00]
coordinates {
(C,0.5324)
(A,0.7549)
(B,0.8077)
(D,0.7769)
};
\addlegendentry{Static training}
\addplot [fill=blue,opacity=1.00]
coordinates {
(C,0.6638)
(A,0.8520)
(B,0.9159)
(D,0.8907)
};
\addlegendentry{Periodic retraining}
\addplot [fill=red,opacity=1.00]
coordinates {
(C,0.6519)
(A,0.8426)
(B,0.9141)
(D,0.8872)
};
\addlegendentry{Drift-aware retraining}
\end{axis}
\end{tikzpicture}
    \caption{Average results per model under each retraining scenario}
    \label{fig:final_res_model}
\end{figure}
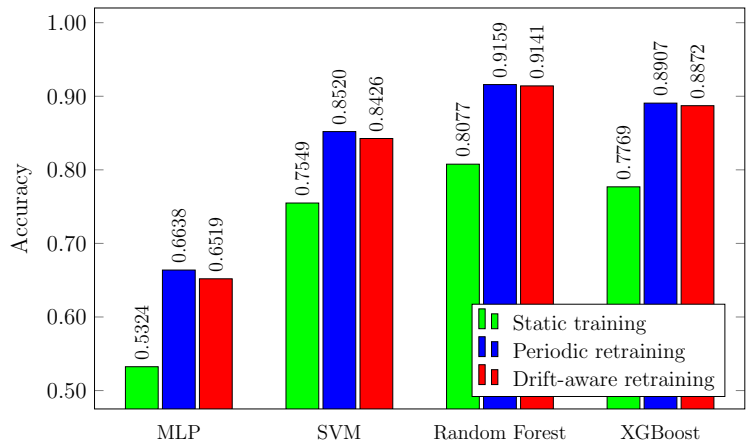

A summary of the average accuracies per scenario for each model is provided in Figure~\ref{fig:final_res_model}. As expected, all of
the models perform relatively poorly in the static scenario,
and much better under the periodic retraining scenario.
For the drift-aware scenario, the accuracy only declines 
slightly---less than~1\%, on average---as compared to periodic retraining, 
which indicates that we are indeed detecting concept drift. 
Given our experimental design, we would not expect that any concept drift 
detection approach would outperform periodic retraining. Furthermore,
given that our batch size is~50 samples, and that we cluster two consecutive
batches, it is clear that we will not always detect concept drift at the precise 
point at which it occurs. From this perspective, the results in 
Figure~\ref{fig:final_res_model} appear to be extremely positive.

In Figure~\ref{fig:final_res_model_per_family} in the Appendix, we further
break down the results in Figure~\ref{fig:final_res_model} per family.
Comparing the static scenario to the other two scenarios in these graphs,
it appears that Malap and SMSreg are the most affected by
concept drift, while Airpush, Agent, and Boxer are less affected.

In Figures~\ref{fig:per_period_A}, \ref{fig:per_period_B}, and~\ref{fig:per_period_C}
in the Appendix, we graph the per-interval accuracy for selected families and models
under the static training, periodic retraining, and drift-aware retraining scenarios, respectively.
We observe that Random Forest and SVM models consistently yield similar results
in each scenario, with XGBoost differing only slightly. The poor performance of  
MLP models is readily apparent in these graphs.

Interestingly, we observe from Figures~\ref{fig:per_period_A}(c) and~(d)
that even in the presence of substantial concept drift, the accuracy
under the static scenario
can improve at some points. While this may initially seem counterintuitive, 
a likely explanation is that some older variants have been recycled,
either with no modification, or with modifications that
do not affect the original model's accuracy. We also note that the
similarities between the two SMSreg graphs 
in each scenario---see the~(c) and~(d) graphs in 
Figures~\ref{fig:per_period_A}, \ref{fig:per_period_B}, 
and~\ref{fig:per_period_C}---serve to further emphasize
the accuracy and consistency of our concept drift detection strategy.

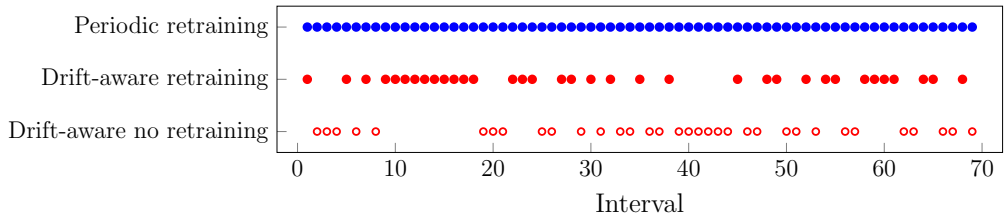
\begin{figure}[!htb]
    \centering
\begin{tikzpicture}[scale=0.85]
    \begin{axis}[
        width=0.835\textwidth,
        height=0.25\textwidth,
        xmin=0, xmax=70,
        ymin=-0.2, ymax=1.2,
        xtick={0,10,20,30,40,50,60,70},
        xticklabels={0,10,20,30,40,50,60,70},
        xtick pos=bottom,
        ytick pos=left,
        x tick label style={scale=0.9},
        y tick label style={scale=0.9},
        ytick={0,0.5,1},
        yticklabels={Drift-aware no retraining, Drift-aware retraining, Periodic retraining},
        enlarge x limits=0.03,
        xlabel={Interval},
     ] 
\addplot[color=blue, very thick, only marks, mark size=1.5pt] coordinates {
(1,1)
(2,1)
(3,1)
(4,1)
(5,1)
(6,1)
(7,1)
(8,1)
(9,1)
(10,1)
(11,1)
(12,1)
(13,1)
(14,1)
(15,1)
(16,1)
(17,1)
(18,1)
(19,1)
(20,1)
(21,1)
(22,1)
(23,1)
(24,1)
(25,1)
(26,1)
(27,1)
(28,1)
(29,1)
(30,1)
(31,1)
(32,1)
(33,1)
(34,1)
(35,1)
(36,1)
(37,1)
(38,1)
(39,1)
(40,1)
(41,1)
(42,1)
(43,1)
(44,1)
(45,1)
(46,1)
(47,1)
(48,1)
(49,1)
(50,1)
(51,1)
(52,1)
(53,1)
(54,1)
(55,1)
(56,1)
(57,1)
(58,1)
(59,1)
(60,1)
(61,1)
(62,1)
(63,1)
(64,1)
(65,1)
(66,1)
(67,1)
(68,1)
(69,1)
};
\addplot[color=red, very thick, only marks, mark size=1.5pt] coordinates {
(1,0.5)
(5,0.5)
(7,0.5)
(9,0.5)
(10,0.5)
(11,0.5)
(12,0.5)
(13,0.5)
(14,0.5)
(15,0.5)
(16,0.5)
(17,0.5)
(18,0.5)
(22,0.5)
(23,0.5)
(24,0.5)
(27,0.5)
(28,0.5)
(30,0.5)
(32,0.5)
(35,0.5)
(38,0.5)
(45,0.5)
(48,0.5)
(49,0.5)
(52,0.5)
(54,0.5)
(55,0.5)
(58,0.5)
(59,0.5)
(60,0.5)
(61,0.5)
(64,0.5)
(65,0.5)
(68,0.5)
};
\addplot[color=red, thick, only marks, mark size=1.5pt, mark=o] coordinates {
(2,0)
(3,0)
(4,0)
(6,0)
(8,0)
(19,0)
(20,0)
(21,0)
(25,0)
(26,0)
(29,0)
(31,0)
(33,0)
(34,0)
(36,0)
(37,0)
(39,0)
(40,0)
(41,0)
(42,0)
(43,0)
(44,0)
(46,0)
(47,0)
(50,0)
(51,0)
(53,0)
(56,0)
(57,0)
(62,0)
(63,0)
(66,0)
(67,0)
(69,0)
};
    \end{axis}
    \end{tikzpicture}
    
    \caption{Retraining intervals for Boxer family}
    \label{fig:retraining_intervals}
\end{figure}

Finally, we consider the savings---in terms of the number of models 
that need to be trained---under the drift-aware scenario, 
as compared to the periodic retraining scenario.
Figure~\ref{fig:retraining_intervals} compares the number of models in the 
periodic retraining and drift-aware retraining scenarios for the Boxer family.
For this family, there are~69 temporal subsets, and hence there 
are~69 models trained in the periodic retraining scenario. 
For the Boxer family, we detect concept drift at~34
time intervals, which implies that for the drift-aware scenario, 
only~35 models are trained, which is a reduction of almost~50\%.
From Figure~\ref{fig:final_res_model_per_family}(d) in the Appendix, 
we see that, on average, this savings in training time
result in only a minimal loss of accuracy---ignoring the 
consistently poor-performing
MLP model, the loss of accuracy is less that~0.5\%, on average.

In Table~\ref{tab:retrain_intervals} we summarize the drift-aware training 
savings for each family. The savings
range from a low of less than~32\%\ for the Malap family, to a high of more 
than~49\%\ for Boxer, while the work reduction 
over all of the families considered is~39.48\%,
with a per-family average reduction of~40.03\%. Thus, our 
concept drift detection technique can be expected to reduce the number of models
that need to be trained by about~40\%, as compared to simply retraining at each
time interval. Furthermore, according to Figure~\ref{fig:final_res_model}, the loss in
accuracy is negligible under this efficient drift-aware scenario, as compared
to periodic retraining.

\begin{table}[!htb]
\def\zz{\phantom{0}}
\centering
\caption{Number of training intervals}\label{tab:retrain_intervals}
\adjustbox{scale=0.85}{
\begin{tabular}{c|ccc}
\toprule
\multirow{2}{*}{\textbf{Family}} & \multicolumn{2}{c}{\textbf{Retraining}} 
		& \multirow{2}{*}{\textbf{Savings}} \\ \cline{2-3} \\[-2.6ex]
	& \textbf{Periodic} & \textbf{Drift-aware}  \\ \midrule
Airpush & 154 & \zz96 & 37.66\% \\ 
SMSreg & \zz84 & \zz50 & 40.48\% \\ 
Malap & \zz70 & \zz48 & 31.43\% \\ 
Boxer & \zz69 & \zz35 & 49.28\% \\ 
Agent & \zz46 & \zz27 & 41.30\% \\ \midrule
Total & 423 & 256 & 39.48\% \\ \bottomrule
\end{tabular}
}
\end{table}

\section{Conclusion}\label{chap:con}

In this paper, we considered the utility of clustering as a means of detecting  
concept drift in the malware domain. We analyzed three different scenarios---static training,
periodic retraining, and drift-aware retraining---and for each of these scenarios,
we trained four different classification models: MLP, SVM, Random Forest, and XGBoost.
In the static training scenario, each learning model was trained on an initial temporal
batch of samples, after which no retraining occurred. This scenario
served as a baseline, and the relatively poor classification results showed 
that at least some degree of concept drift occurs in each of
the malware families tested, although the extent of the drift varied significantly
for different families.

In the periodic retraining scenario, we retrained each model whenever a new 
temporal batch of samples had been accumulated since the previous retraining.
Our results for this scenario demonstrate that we can greatly improve on the 
accuracy of the static training scenario by frequently retraining learning models. 
However, this periodic retraining approach is computationally costly, as models 
need to be constantly retrained. 

Under the drift-aware retraining scenario we only retrained models when
concept drift was detected. To detect concept drift, we employed MiniBatch $K$-Means
clustering and computed the average silhouette coefficient, with an empirically-determined
threshold serving to trigger model retraining. We found that this drift-aware retraining
approach yielded results that were only marginally worse than the periodic retraining scenario, while
being far more computationally efficient, due to less frequent retraining.
These results serve to verify that our clustering-based concept drift detection technique 
provides a practical and efficient method for 
improving the accuracy of malware classification models.

There are several potentially fruitful paths for future research. For example,
the analysis of the three retraining scenarios  
considered in this paper could be expanded to include additional malware families and 
learning techniques. In addition, our clustering-based detection strategy 
could be directly compared to---and possibly combined with---the machine 
learning based techniques in the related series of papers~\cite{Sunhera,Lolitha,Mayuri}.

As another avenue of future work, our clustering-based drift detection approach
could likely be improved by applying a more 
sophisticated analysis to the silhouette coefficients. In a similar vein, additional
thresholding work would likely improve on the results presented in this paper. It would also
be interesting to determine whether other clustering techniques---such as 
EM clustering and density-based clustering---could 
improve on the results presented in this paper.

\bibliographystyle{plain}
\bibliography{references}

\begin{thebibliography}{10}

\bibitem{Anupama2022}
M.~L. Anupama, P.~Vinod, Corrado~Aaron Visaggio, M.~A. Arya, Josna Philomina,
  Rincy Raphael, Anson Pinhero, K.~S. Ajith, and P.~Mathiyalagan.
\newblock Detection and robustness evaluation of {A}ndroid malware classifiers.
\newblock {\em Journal of Computer Virology and Hacking Techniques},
  18(3):147--170, 2022.

\bibitem{bm}
Jean-Marie Borello and Ludovic M\'{e}.
\newblock Code obfuscation techniques for metamorphic viruses.
\newblock {\em Journal in Computer Virology}, 4(3):211--220, 2008.

\bibitem{Yizheng}
Yizheng Chen, Zhoujie Ding, and David Wagner.
\newblock Continuous learning for android malware detection.
\newblock In {\em 32nd {USENIX} Security Symposium}, USENIX Security 23, pages
  1127--1144, 2023.

\bibitem{Chen}
Zhi Chen, Zhenning Zhang, Zeliang Kan, Limin Yang, Jacopo Cortellazzi, Feargus
  Pendlebury, Fabio Pierazzi, Lorenzo Cavallaro, and Gang Wang.
\newblock Is it overkill? {A}nalyzing feature-space concept drift in malware
  detectors.
\newblock In {\em 2023 IEEE Security and Privacy Workshops}, SPW, pages 21--28,
  2023.

\bibitem{Agent}
{F-Secure: Android/Agent}.
\newblock \url{https://www.f-secure.com/v-descs/agent.shtml}, 2024.

\bibitem{AirPush}
{F-Secure: Android/AirPush}.
\newblock \url{https://www.f-secure.com/v-descs/trojan-android-airpush.shtml},
  2024.

\bibitem{Boxer}
{F-Secure: Android/Boxer}.
\newblock \url{https://www.f-secure.com/v-descs/trojan-android-boxer.shtml},
  2024.

\bibitem{SMSReg}
{F-Secure: Android/SMSReg}.
\newblock \url{https://www.f-secure.com/sw-desc/riskware-android-smsreg.shtml},
  2024.

\bibitem{towardsdatascience_mlp}
Aly~El Gamal.
\newblock Multilayer perceptron explained with a real-life example and python
  code: Sentiment analysis.
\newblock
  \url{https://towardsdatascience.com/multilayer-perceptron-explained-with-a-real-life-example-and-python-code-sentiment-analysis-cb408ee93141},
  2020.

\bibitem{Gibert}
Daniel Gibert, Carles Mateu, and Jordi Planes.
\newblock The rise of machine learning for detection and classification of
  malware: Research developments, trends and challenges.
\newblock {\em Journal of Network and Computer Applications}, 153:102526, 2020.

\bibitem{Krono}
Alejandro Guerra-Manzanares, Hayretdin Bahsi, and Sven Nõmm.
\newblock Kronodroid: Time-based hybrid-featured dataset for effective
  {A}ndroid malware detection and characterization.
\newblock {\em Computers \&\ Security}, 110:102399, 2021.

\bibitem{Alejandro}
Alejandro Guerra-Manzanares, Marcin Luckner, and Hayretdin Bahsi.
\newblock Concept drift and cross-device behavior: Challenges and implications
  for effective {A}ndroid malware detection.
\newblock {\em Computers \& Security}, 120:102757, 2022.

\bibitem{Hinckley}
David~V. Hinkley.
\newblock Inference about the change-point in a sequence of random variables.
\newblock {\em Biometrika}, 57(1):1--17, 1970.

\bibitem{Jog}
Anita Jog and Anjali~A. Chandavale.
\newblock Implementation of credit card fraud detection system with concept
  drifts adaptation.
\newblock In Subhash Bhalla, Vikrant Bhateja, Anjali~A. Chandavale, Anil~S.
  Hiwale, and Suresh~Chandra Satapathy, editors, {\em Intelligent Computing and
  Information and Communication}, pages 467--477, 2018.

\bibitem{Kan}
Z.~Kan, F.~Pendlebury, F.~Pierazzi, and L.~Cavallaro.
\newblock Investigating labelless drift adaptation for malware detection.
\newblock In {\em Proceedings of the 14th {ACM} Workshop on Artificial
  Intelligence and Security}, AISec '21, pages 123--134, 2021.

\bibitem{Karbab}
ElMouatez~Billah Karbab and Mourad Debbabi.
\newblock {PetaDroid}: Adaptive {A}ndroid malware detection using deep
  learning.
\newblock In {\em Detection of Intrusions and Malware, and Vulnerability
  Assessment: 18th International Conference}, DIMVA, pages 319--340, 2021.

\bibitem{ibm_xgboost}
Eda Kavlakoglu and Erika Russi.
\newblock What is {XGBoost}?
\newblock \url{https://www.ibm.com/topics/xgboost}, 2024.

\bibitem{Lin}
Da~Lin and Mark Stamp.
\newblock Hunting for undetectable metamorphic viruses.
\newblock {\em Journal in Computer Virology}, 7(3):201--214, 2011.

\bibitem{Lout2024}
Pavla Louthánová, Matouš Kozák, Martin Jureček, Mark Stamp, and Fabio~Di
  Troia.
\newblock A comparison of adversarial malware generators.
\newblock {\em Journal of Computer Virology and Hacking Techniques},
  20(4):623--639, 2024.

\bibitem{luomaaho2023analysis}
Mika Luoma-aho.
\newblock Analysis of modern malware: Obfuscation techniques.
\newblock \url{http://www.theseus.fi/handle/10024/798038}, 2023.

\bibitem{github}
Alejandro~Guerra Manzanares.
\newblock About the {K}ronodroid dataset.
\newblock \url{https://github.com/aleguma/kronodroid/blob/main/README.md},
  2024.

\bibitem{Molina}
Borja Molina-Coronado, Usue Mori, Alexander Mendiburu, and Jose Miguel-Alonso.
\newblock Efficient concept drift handling for batch {A}ndroid malware
  detection models.
\newblock {\em Pervasive and Mobile Computing}, 96:101849, 2023.

\bibitem{Page}
Ewan~S. Page.
\newblock Continuous inspection schemes.
\newblock {\em Biometrika}, 41(1/2):100--115, 1954.

\bibitem{Sunhera}
Sunhera Paul and Mark Stamp.
\newblock Word embedding techniques for malware evolution detection.
\newblock In Mark Stamp, Mamoun Alazab, and Andrii Shalaginov, editors, {\em
  Malware Analysis Using Artificial Intelligence and Deep Learning}, pages
  321--343. Springer, 2021.

\bibitem{minibatch_k_means}
Mithilesh Pradhan.
\newblock Tutorialspoint: Mini batch $k$-means clustering algorithm in machine
  learning.
\newblock
  \url{https://www.tutorialspoint.com/mini-batch-k-means-clustering-algorithm-in-machine-learning},
  2023.

\bibitem{scikit-learn_mini_batch_kmeans}
Scikit-learn developers: sklearn.cluster.minibatchkmeans.
\newblock
  \url{https://scikit-learn.org/dev/modules/generated/sklearn.cluster.MiniBatchKMeans.html},
  2024.

\bibitem{scikit-learn_random_forest_classifier}
Scikit-learn developers: sklearn.ensemble.randomforestclassifier.
\newblock
  \url{https://scikit-learn.org/stable/modules/generated/sklearn.ensemble.RandomForestClassifier.html},
  2024.

\bibitem{scikit-learn_mlp_classifier}
Scikit-learn developers: sklearn.neural\_network.mlpclassifier.
\newblock
  \url{https://scikit-learn.org/dev/modules/generated/sklearn.neural\_network.MLPClassifier.html},
  2024.

\bibitem{scikit-learn_linear_svc}
Scikit-learn developers: sklearn.svm.linearsvc.
\newblock
  \url{https://scikit-learn.org/dev/modules/generated/sklearn.svm.LinearSVC.html},
  2024.

\bibitem{Sharma_2014}
Ashu Sharma and S.~K.~Sahay.
\newblock Evolution and detection of polymorphic and metamorphic malwares: A
  survey.
\newblock {\em International Journal of Computer Applications}, 90(2):7--11,
  2014.

\bibitem{10.1145/2381896.2381910}
Anshuman Singh, Andrew Walenstein, and Arun Lakhotia.
\newblock Tracking concept drift in malware families.
\newblock In {\em Proceedings of the 5th ACM Workshop on Security and
  Artificial Intelligence}, AISec '12, pages 81--92, 2012.

\bibitem{9027485}
Siddharth Singhal, Utkarsh Chawla, and Rajeev Shorey.
\newblock Machine learning \& concept drift based approach for malicious
  website detection.
\newblock In {\em 2020 International Conference on Communication Systems \&
  Networks}, COMSNETS, pages 582--585, 2020.

\bibitem{ibm_svm}
{S}upport {V}ector {M}achine.
\newblock \url{https://www.ibm.com/topics/support-vector-machine}, 2023.

\bibitem{Lolitha}
Lolitha~Sresta Tupadha and Mark Stamp.
\newblock Machine learning for malware evolution detection.
\newblock In Mark Stamp, Corrado Aaron~Visaggio, Francesco Mercaldo, and Fabio
  Di~Troia, editors, {\em Artificial Intelligence for Cybersecurity}, pages
  183--213. Springer, 2022.

\bibitem{Mayuri}
Mayuri Wadkar, Fabio {Di Troia}, and Mark Stamp.
\newblock Detecting malware evolution using {S}upport {V}ector {M}achines.
\newblock {\em Expert Systems with Applications}, 143:113022, 2020.

\bibitem{ibm_random_forest}
What is random forest?
\newblock \url{https://www.ibm.com/topics/random-forest}.

\bibitem{xgboost_sklearn_estimator}
{XGBoost} developers: {XGBoost} {Python} sklearn estimator.
\newblock
  \url{https://xgboost.readthedocs.io/en/latest/python/sklearn_estimator.html},
  2024.

\bibitem{Xu}
Ke~Xu, Yingjiu Li, Robert Deng, Kai Chen, and Jiayun Xu.
\newblock {DroidEvolver}: Self-evolving {A}ndroid malware detection system.
\newblock In {\em 2019 IEEE European Symposium on Security and Privacy},
  EuroS\&P, pages 47--62, 2019.

\end{thebibliography}

\section*{Appendix}\label{app:a}

\titleformat{\section}{\normalfont\large\bfseries}{}{0em}{#1\ \thesection}
\setcounter{section}{0}
\renewcommand{\thesection}{\Alph{section}}
\renewcommand{\thesubsection}{A.\arabic{subsection}}
\setcounter{table}{0}
\renewcommand{\thetable}{A.\arabic{table}}
\setcounter{figure}{0}
\renewcommand{\thefigure}{A.\arabic{figure}}

\subsection{Additional Silhouette Coefficient Graphs}

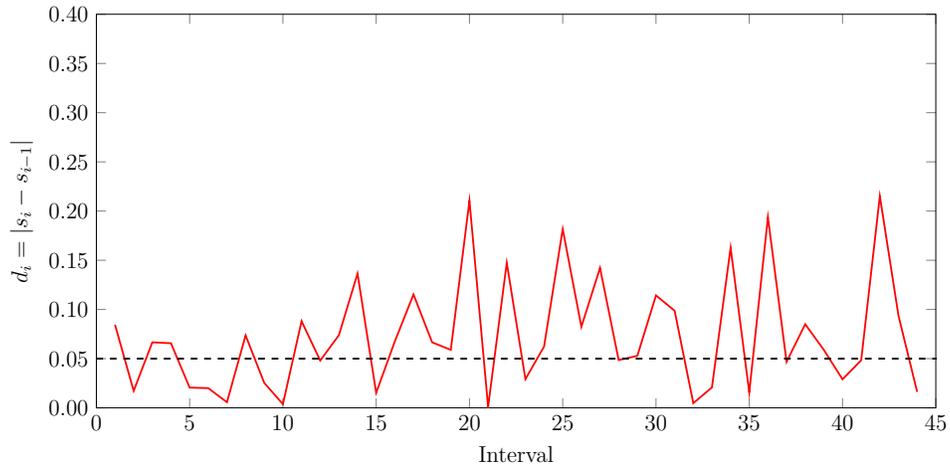
\begin{figure}[!htb]
    \centering
    \begin{tikzpicture}[scale=0.85]
\begin{axis}[xmin=0, xmax=45, ymin=0.0, ymax=0.40,
		   width=0.95\textwidth,
		   height=0.50\textwidth,
	 	   x tick label style={scale=0.85,
   		 	/pgf/number format/.cd,
			/pgf/number format/1000 sep={},
   			fixed,
   			fixed zerofill,
    			precision=0
		   },
		   x label style={scale=0.85},
	 	   y tick label style={scale=0.85,
    		 	/pgf/number format/.cd,
   			fixed,
   			fixed zerofill,
    			precision=2
		    },
		   y label style={scale=0.85},
                    xtick={0,5,10,15,20,25,30,35,40,45},
                    ytick={0.0,0.05,0.1,0.15,0.2,0.25,0.3,0.35,0.40},
                    xlabel={Interval},
                    ylabel={$d_i = |s_i-s_{i-1}|$}] 
\addplot[color=red, thick, mark=none] coordinates { 
(1, 0.08444932991838866)
(2, 0.017244635358385288)
(3, 0.06663188403254258)
(4, 0.0656645498181744)
(5, 0.020587702352622445)
(6, 0.020028721861345126)
(7, 0.005746934687309288)
(8, 0.0736253744525725)
(9, 0.025331438063129125)
(10, 0.003791172511824481)
(11, 0.08809821339024332)
(12, 0.048067145690455004)
(13, 0.07384804226842231)
(14, 0.13645380873561216)
(15, 0.01524639828345764)
(16, 0.06754730988223262)
(17, 0.11526177812950392)
(18, 0.0665959944722729)
(19, 0.05892379060161926)
(20, 0.2105647836573259)
(21, 0.0009016552825239499)
(22, 0.1474435902657027)
(23, 0.02920499612832951)
(24, 0.06228474364853098)
(25, 0.18162425697779)
(26, 0.08238794029117734)
(27, 0.14218327095258126)
(28, 0.04853900274894696)
(29, 0.05286670855113851)
(30, 0.1142940994526816)
(31, 0.09864557259897877)
(32, 0.004739534869654671)
(33, 0.020780367865231464)
(34, 0.16270599888396634)
(35, 0.015775904922295497)
(36, 0.1936886025274514)
(37, 0.04678265418128169)
(38, 0.08503293318329735)
(39, 0.05884695918042404)
(40, 0.029024463053977545)
(41, 0.04837676112101863)
(42, 0.21516514841199869)
(43, 0.09363792553478423)
(44, 0.016243071976532686)
};
\addplot[color=black, thick, dashed] table[row sep = crcr]{0 0.05 \\ 152 0.05 \\};
\end{axis}
\end{tikzpicture}
    \caption{Silhouette score differences for Agent}\label{fig:agent_sil}
\end{figure}

\begin{figure}[!htb]
    \centering
    \begin{tikzpicture}[scale=0.85]
\begin{axis}[xmin=0, xmax=153, ymin=0.0, ymax=0.40,
		   width=0.95\textwidth,
		   height=0.50\textwidth,
	 	   x tick label style={scale=0.85,
   		 	/pgf/number format/.cd,
			/pgf/number format/1000 sep={},
   			fixed,
   			fixed zerofill,
    			precision=0
		   },
		   x label style={scale=0.85},
	 	   y tick label style={scale=0.85,
    		 	/pgf/number format/.cd,
   			fixed,
   			fixed zerofill,
    			precision=2
		    },
		   y label style={scale=0.85},
                    xtick={0,10,20,30,40,50,60,70,80,90,100,110,120,130,140,150},
                    ytick={0.0,0.05,0.1,0.15,0.2,0.25,0.3,0.35,0.40},
                    xlabel={Interval},
                    ylabel={$d_i = |s_i-s_{i-1}|$}] 
\addplot[color=red, thick, mark=none] coordinates { 
(1, 0.01743528402261589)
(2, 0.13373077102878034)%
(3, 0.027012077425515008)
(4, 0.07382808657997819)%
(5, 0.21852390152528353)%
(6, 0.11764203892385178)%
(7, 0.026395340816235097)
(8, 0.04216024487645065)
(9, 0.04964862014065366)
(10, 0.04581763426731142)
(11, 0.07753883257896926)%
(12, 0.1304020917954295)%
(13, 0.19373103166632877)%
(14, 0.23334303791803462)%
(15, 0.03303768079776079)
(16, 0.09841305033359624)%
(17, 0.009150940625636927)
(18, 0.03577248509165776)
(19, 0.13624939059772012)%
(20, 0.029951726130680978)
(21, 0.08639503608384025)%
(22, 0.062218906525590756)%
(23, 0.20849932944748895)%
(24, 0.17103383299275118)%
(25, 0.05333658458595614)%
(26, 0.11341634423483318)%
(27, 0.05181274573576222)%
(28, 0.05704167085266157)%
(29, 0.04556098840431605)
(30, 0.17487639384950607)%
(31, 0.23552697666545755)%
(32, 0.11315591895831131)%
(33, 0.05615323027948087)%
(34, 0.1393708285891777)%
(35, 0.034599656414753335)
(36, 0.11800563249188685)%
(37, 0.14846995323320356)%
(38, 0.07100446568967816)%
(39, 0.05650725911152338)%
(40, 0.09341342947421943)%
(41, 0.005202811264950036)
(42, 0.019532307681486366)
(43, 0.0883007958591249)%
(44, 0.01056726020784532)
(45, 0.01188745262538049)
(46, 0.15503804357792628)%
(47, 0.051816075553795526)%
(48, 0.005613395759054451)
(49, 0.041040243902571716)
(50, 0.15321145465562896)%
(51, 0.13109532517470282)%
(52, 0.10317536352552889)%
(53, 0.09664301265379166)%
(54, 0.16399627040217563)%
(55, 0.0025884249419322533)
(56, 0.10772094143320449)%
(57, 0.026311043816498847)
(58, 0.08057134068285793)%
(59, 0.01271577585636624)
(60, 0.17372955589504693)%
(61, 0.008317418880705016)
(62, 0.026628435316729968)
(63, 0.0747018286856811)%
(64, 0.1428859644621294)%
(65, 0.12161512702065193)%
(66, 0.038056853375573255)
(67, 0.09717530944068783)%
(68, 0.11400648344698563)%
(69, 0.026178298503828135)
(70, 0.0015926282885295184)
(71, 0.05258017122680689)%
(72, 0.007645548287384368)
(73, 0.13210310485799512)%
(74, 0.02812246544271113)
(75, 0.29006769291940415)%
(76, 0.21435314135509387)%
(77, 0.024970940271326292)
(78, 0.043065797624217)
(79, 0.08302569385865996)%
(80, 0.16555946231472696)%
(81, 0.1777141106872298)%
(82, 0.021381208065680224)
(83, 0.016639891391438455)
(84, 0.01748048776475164)
(85, 0.21109794116689656)%
(86, 0.055753041202876004)%
(87, 0.1861746180111084)%
(88, 0.003823056422360749)
(89, 0.03519783951807931)
(90, 0.12282565335510365)%
(91, 0.0753855225126781)%
(92, 0.007848226734326974)
(93, 0.014424796890814162)
(94, 0.06464907696677136)%
(95, 0.07548870363150534)%
(96, 0.054287743986308715)%
(97, 0.20405278965074553)%
(98, 0.01862384980121931)
(99, 0.06207416994675424)%
(100, 0.05686692813089961)%
(101, 0.03323821140235972)
(102, 0.06131297922012813)%
(103, 0.02622795460016497)
(104, 0.02451482489749246)
(105, 0.06334086256539018)%
(106, 0.010167325045174813)
(107, 0.017831777688085748)
(108, 0.08772996310501494)%
(109, 0.04795885989750573)
(110, 0.06346643762294915)%
(111, 0.14082901186247046)%
(112, 0.07379897473186192)%
(113, 0.05092561917909336)%
(114, 0.12114730698949988)%
(115, 0.11414816071905845)%
(116, 0.03913382837307944)
(117, 0.04987920442028823)
(118, 0.1456688106253554)%
(119, 0.1100796448759809)%
(120, 0.10455485186493982)%
(121, 0.1026820801346866)%
(122, 0.0740303777199295)%
(123, 0.18289869559279426)%
(124, 0.1257156866604306)%
(125, 0.08781641623287217)%
(126, 0.07487922132659475)%
(127, 0.08722884612823822)%
(128, 0.27453712431732574)%
(129, 0.19666392968752333)%
(130, 0.0389611229420595)
(131, 0.04586924521034308)
(132, 0.0283932026951563)
(133, 0.06049424404378906)%
(134, 0.06371828047994896)%
(135, 0.31692322985317745)%
(136, 0.3348924546718697)%
(137, 0.017572925166031678)
(138, 0.06383940073075645)%
(139, 0.013323768703220634)
(140, 0.021013489548163733)
(141, 0.05438017741655721)%
(142, 0.029440507059259102)
(143, 0.11753537895501415)%
(144, 0.06663596125248478)%
(145, 0.03116904092927375)
(146, 0.15356560064551633)%
(147, 0.06619521231458012)%
(148, 0.08638319875883374)%
(149, 0.08385980146936733)%
(150, 0.03889037855417282)
(151, 0.04331867359082958)
(152, 0.01506041588130394)
};
\addplot[color=black, thick, dashed] table[row sep = crcr]{0 0.05 \\ 152 0.05 \\};
\end{axis}
\end{tikzpicture}
    \caption{Silhouette score differences for Airpush}\label{fig:airpush_sil}
\end{figure}
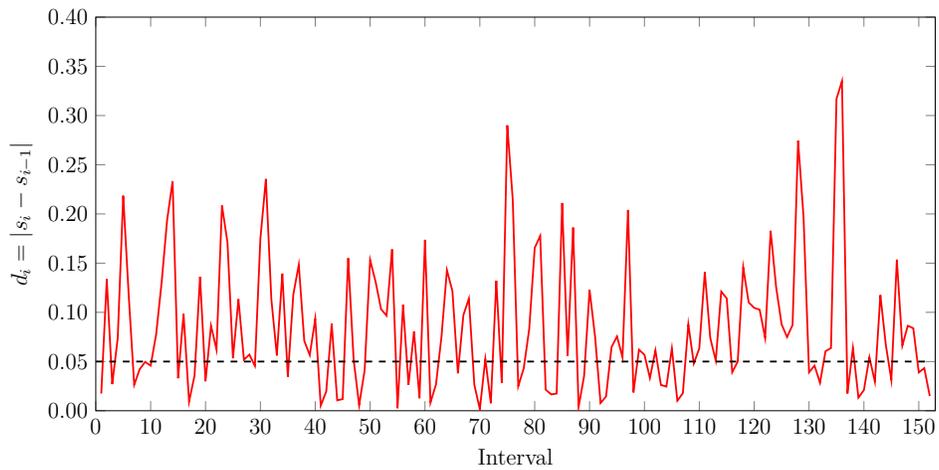

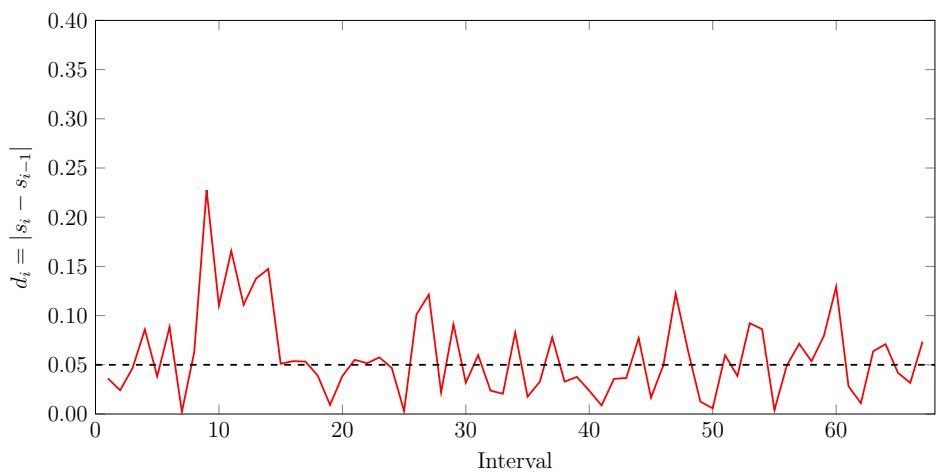
\begin{figure}[!htb]
    \centering
    \begin{tikzpicture}[scale=0.85]
\begin{axis}[xmin=0, xmax=68, ymin=0.0, ymax=0.40,
		   width=0.95\textwidth,
		   height=0.50\textwidth,
	 	   x tick label style={scale=0.85,
   		 	/pgf/number format/.cd,
			/pgf/number format/1000 sep={},
   			fixed,
   			fixed zerofill,
    			precision=0
		   },
		   x label style={scale=0.85},
	 	   y tick label style={scale=0.85,
    		 	/pgf/number format/.cd,
   			fixed,
   			fixed zerofill,
    			precision=2
		    },
		   y label style={scale=0.85},
                    xtick={0,10,20,30,40,50,60},
                    ytick={0.0,0.05,0.1,0.15,0.2,0.25,0.3,0.35,0.40},
                    xlabel={Interval},
                    ylabel={$d_i = |s_i-s_{i-1}|$}] 
\addplot[color=red, thick, mark=none] coordinates { 
(1, 0.03615558604420649)
(2, 0.024151956913506245)
(3, 0.04664215951759759)
(4, 0.08585946049454052)
(5, 0.038290756914279966)
(6, 0.08827092542499126)
(7, 0.0021808263887057033)
(8, 0.06305800621459351)
(9, 0.22756579942241908)
(10, 0.11013800209800687)
(11, 0.16571118894325593)
(12, 0.11105729306394868)
(13, 0.13749725200951296)
(14, 0.1475253556895444)
(15, 0.05109446411769469)
(16, 0.05371930810712644)
(17, 0.053311749052119495)
(18, 0.039166742747811456)
(19, 0.009223993273713593)
(20, 0.03858917275065904)
(21, 0.0550443082006522)
(22, 0.051637018304355964)
(23, 0.05751780870385409)
(24, 0.04681940374291843)
(25, 0.003058384927647484)
(26, 0.1011196289128061)
(27, 0.12132979010259207)
(28, 0.02199754465895637)
(29, 0.09106919812249492)
(30, 0.03186346557658848)
(31, 0.05986840724528486)
(32, 0.02374218769607145)
(33, 0.02057613624242982)
(34, 0.08270337945661319)
(35, 0.0176159934185314)
(36, 0.0329002531424728)
(37, 0.07786970988707698)
(38, 0.03291476852002678)
(39, 0.03769338244174203)
(40, 0.023655168441908625)
(41, 0.008722818615576389)
(42, 0.03572347498823364)
(43, 0.036595278000759346)
(44, 0.07707713734798194)
(45, 0.01661860142813576)
(46, 0.04947469641781824)
(47, 0.12236667857485525)
(48, 0.06424722185680354)
(49, 0.012512640215757875)
(50, 0.005766115450072662)
(51, 0.05981551867494794)
(52, 0.03871341652418153)
(53, 0.09223320892325465)
(54, 0.08626808424440363)
(55, 0.0040758828554510185)
(56, 0.04920482694603362)
(57, 0.07144235314263875)
(58, 0.05367288973854817)
(59, 0.07922172346975018)
(60, 0.1292924813603444)
(61, 0.028216463409640125)
(62, 0.01102390457307606)
(63, 0.06377412115375902)
(64, 0.07098839234035631)
(65, 0.04194669932261971)
(66, 0.031599198098817255)
(67, 0.0735797236765886)
};
\addplot[color=black, thick, dashed] table[row sep = crcr]{0 0.05 \\ 152 0.05 \\};
\end{axis}
\end{tikzpicture}
    \caption{Silhouette score differences for Boxer}\label{fig:boxer_sil}
\end{figure}

In this section, we provide silhouette coefficient graphs for the Agent, Airpush, and Boxer families
in Figures~\ref{fig:agent_sil}, \ref{fig:airpush_sil}, and~\ref{fig:boxer_sil}, respectively.
The silhouette coefficient graphs for the other two families, SMSreg and Malap,
appear in Section~\ref{sect:setup} is Figures~\ref{fig:SMS_sil} and~\ref{fig:malap_sil}, respectively.

%
%


\clearpage

\subsection{Drift Results per Family}

Here, we break down the results in Figure~\ref{fig:final_res_model} per family.
These graphs are discuss in Section~\ref{sect:disc}.

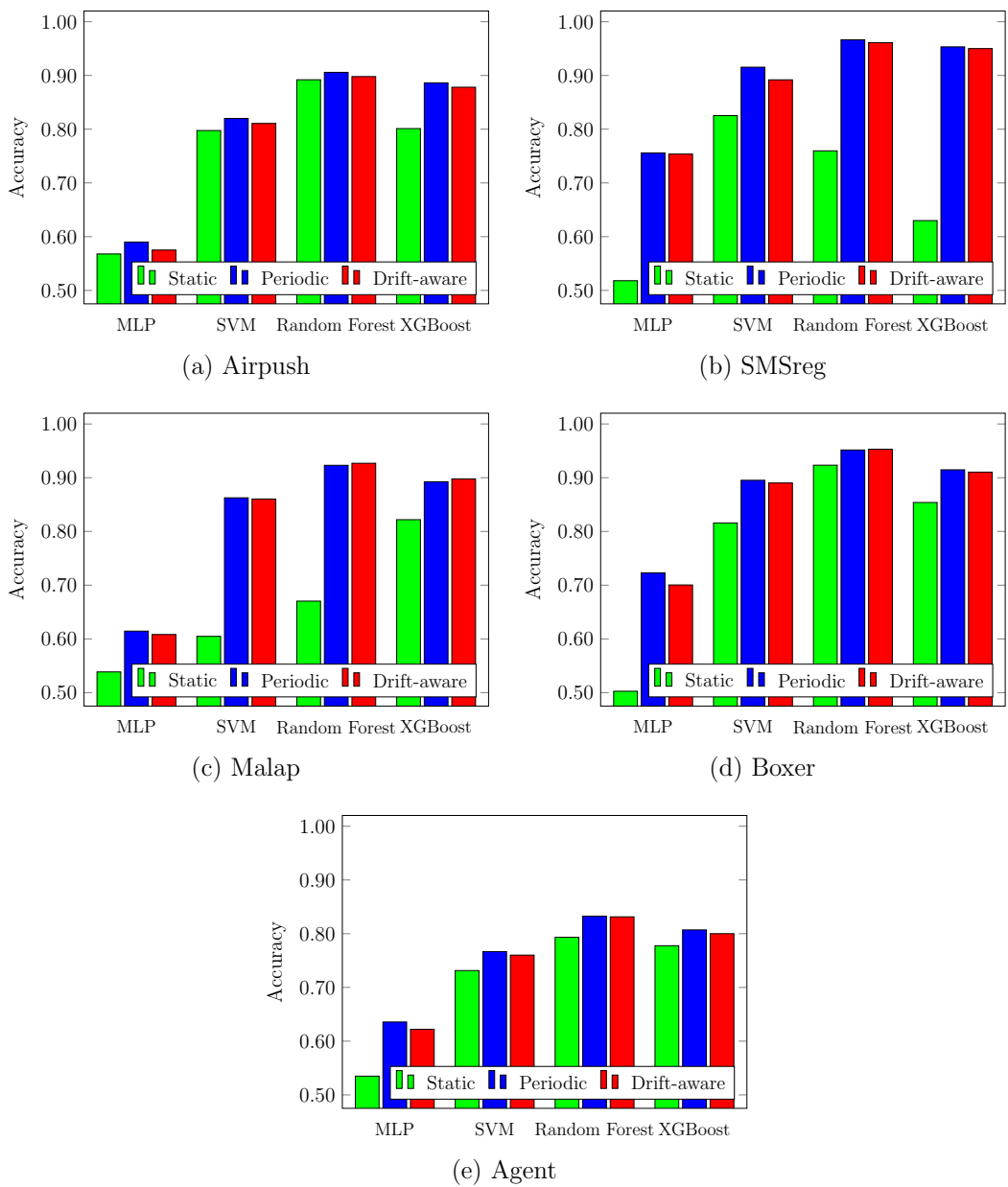
\begin{figure}[!htb]
    \centering
    \begin{tabular}{ccc}
    \begin{tikzpicture}[scale=0.75, every node/.style={scale=1.0}]
\pgfkeys{/pgf/number format/.cd,1000 sep={}}
\begin{axis}[
        width  = 0.585*\textwidth,
        height = 7.0cm,
        ymin=0.475,ymax=1.02,
        ytick={0.50, 0.60, 0.70, 0.80, 0.90, 1.00},
        major x tick style = transparent,
        ybar=5*\pgflinewidth,
        bar width=12.5pt,
        ylabel = {Accuracy},
        ylabel style = {scale = 0.9},
        symbolic x coords={C, A, B, D},
        xticklabels={MLP,  SVM, Random Forest, XGBoost},
 	y tick label style={scale=0.9,
    		/pgf/number format/.cd,
   		fixed,
   		fixed zerofill,
    		precision=2},
        xtick = data,
        x tick label style={scale=0.8,
		},
        enlarge x limits=0.175,
        legend cell align=left,
        legend pos=south east,
        legend columns=3,
        legend style={nodes={scale=0.85},
                column sep=1ex
        },
]
\addplot [fill=green,opacity=1.00]
coordinates {
(C,0.5678)
(A,0.7975)
(B,0.8920)
(D,0.8011)
};
\addlegendentry{Static}
\addplot [fill=blue,opacity=1.00]
coordinates {
(C,0.5900)
(A,0.8200)
(B,0.9058)
(D,0.8863)
};
\addlegendentry{Periodic}
\addplot [fill=red,opacity=1.00]
coordinates {
(C,0.5751)
(A,0.8107)
(B,0.8979)
(D,0.8781)
};
\addlegendentry{Drift-aware}
\end{axis}
\end{tikzpicture}
    &
    \begin{tikzpicture}[scale=0.75, every node/.style={scale=1.0}]
\pgfkeys{/pgf/number format/.cd,1000 sep={}}
\begin{axis}[
        width  = 0.585*\textwidth,
        height = 7.0cm,
        ymin=0.475,ymax=1.02,
        ytick={0.50, 0.60, 0.70, 0.80, 0.90, 1.00},
        major x tick style = transparent,
        ybar=5*\pgflinewidth,
        bar width=12.5pt,
        ylabel = {Accuracy},
        ylabel style = {scale = 0.9},
        symbolic x coords={C, A, B, D},
        xticklabels={MLP,  SVM, Random Forest, XGBoost},
 	y tick label style={scale=0.9,
    		/pgf/number format/.cd,
   		fixed,
   		fixed zerofill,
    		precision=2},
        xtick = data,
        x tick label style={scale=0.8,
		},
        enlarge x limits=0.175,
        legend cell align=left,
        legend pos=south east,
        legend columns=3,
        legend style={nodes={scale=0.85},
                column sep=1ex
        },
]
\addplot [fill=green,opacity=1.00]
coordinates {
(C,0.5180)
(A,0.8253)
(B,0.7597)
(D,0.6299)
};
\addlegendentry{Static}
\addplot [fill=blue,opacity=1.00]
coordinates {
(C,0.7557)
(A,0.9154)
(B,0.9663)
(D,0.9532)
};
\addlegendentry{Periodic}
\addplot [fill=red,opacity=1.00]
coordinates {
(C,0.7537)
(A,0.8919)
(B,0.9611)
(D,0.9500)
};
\addlegendentry{Drift-aware}
\end{axis}
\end{tikzpicture}
    \\
    \adjustbox{scale=0.85}{(a) Airpush}
    &
    \adjustbox{scale=0.85}{(b) SMSreg}
    \\ \\[-1.25ex]
    \begin{tikzpicture}[scale=0.75, every node/.style={scale=1.0}]
\pgfkeys{/pgf/number format/.cd,1000 sep={}}
\begin{axis}[
        width  = 0.585*\textwidth,
        height = 7.0cm,
        ymin=0.475,ymax=1.02,
        ytick={0.50, 0.60, 0.70, 0.80, 0.90, 1.00},
        major x tick style = transparent,
        ybar=5*\pgflinewidth,
        bar width=12.5pt,
        ylabel = {Accuracy},
        ylabel style = {scale = 0.9},
        symbolic x coords={C, A, B, D},
        xticklabels={MLP,  SVM, Random Forest, XGBoost},
 	y tick label style={scale=0.9,
    		/pgf/number format/.cd,
   		fixed,
   		fixed zerofill,
    		precision=2},
        xtick = data,
        x tick label style={scale=0.8,
		},
        enlarge x limits=0.175,
        legend cell align=left,
        legend pos=south east,
        legend columns=3,
        legend style={nodes={scale=0.85},
                column sep=1ex
        },
]
\addplot [fill=green,opacity=1.00]
coordinates {
(C,0.5386)
(A,0.6048)
(B,0.6703)
(D,0.8219)
};
\addlegendentry{Static}
\addplot [fill=blue,opacity=1.00]
coordinates {
(C,0.6145)
(A,0.8626)
(B,0.9231)
(D,0.8924)
};
\addlegendentry{Periodic}
\addplot [fill=red,opacity=1.00]
coordinates {
(C,0.6082)
(A,0.8603)
(B,0.9269)
(D,0.8976)
};
\addlegendentry{Drift-aware}
\end{axis}
\end{tikzpicture}
    &
    \begin{tikzpicture}[scale=0.75, every node/.style={scale=1.0}]
\pgfkeys{/pgf/number format/.cd,1000 sep={}}
\begin{axis}[
        width  = 0.585*\textwidth,
        height = 7.0cm,
        ymin=0.475,ymax=1.02,
        ytick={0.50, 0.60, 0.70, 0.80, 0.90, 1.00},
        major x tick style = transparent,
        ybar=5*\pgflinewidth,
        bar width=12.5pt,
        ylabel = {Accuracy},
        ylabel style = {scale = 0.9},
        symbolic x coords={C, A, B, D},
        xticklabels={MLP,  SVM, Random Forest, XGBoost},
 	y tick label style={scale=0.9,
    		/pgf/number format/.cd,
   		fixed,
   		fixed zerofill,
    		precision=2},
        xtick = data,
        x tick label style={scale=0.8,
		},
        enlarge x limits=0.175,
        legend cell align=left,
        legend pos=south east,
        legend columns=3,
        legend style={nodes={scale=0.85},
                column sep=1ex
        },
]
\addplot [fill=green,opacity=1.00]
coordinates {
(C,0.5026)
(A,0.8158)
(B,0.9233)
(D,0.8539)
};
\addlegendentry{Static}
\addplot [fill=blue,opacity=1.00]
coordinates {
(C,0.7229)
(A,0.8952)
(B,0.9516)
(D,0.9147)
};
\addlegendentry{Periodic}
\addplot [fill=red,opacity=1.00]
coordinates {
(C,0.7004)
(A,0.8902)
(B,0.9530)
(D,0.9104)
};
\addlegendentry{Drift-aware}
\end{axis}
\end{tikzpicture}
    \\
    \adjustbox{scale=0.85}{(c) Malap}
    &
    \adjustbox{scale=0.85}{(d) Boxer}
    \\ \\[-1.25ex]
    \multicolumn{2}{c}{\begin{tikzpicture}[scale=0.75, every node/.style={scale=1.0}]
\pgfkeys{/pgf/number format/.cd,1000 sep={}}
\begin{axis}[
        width  = 0.585*\textwidth,
        height = 7.0cm,
        ymin=0.475,ymax=1.02,
        ytick={0.50, 0.60, 0.70, 0.80, 0.90, 1.00},
        major x tick style = transparent,
        ybar=5*\pgflinewidth,
        bar width=12.5pt,
        ylabel = {Accuracy},
        ylabel style = {scale = 0.9},
        symbolic x coords={C, A, B, D},
        xticklabels={MLP,  SVM, Random Forest, XGBoost},
 	y tick label style={scale=0.9,
    		/pgf/number format/.cd,
   		fixed,
   		fixed zerofill,
    		precision=2},
        xtick = data,
        x tick label style={scale=0.8,
		},
        enlarge x limits=0.175,
        legend cell align=left,
        legend pos=south east,
        legend columns=3,
        legend style={nodes={scale=0.85},
                column sep=1ex
        },
]
\addplot [fill=green,opacity=1.00]
coordinates {
(C,0.5349)
(A,0.7314)
(B,0.7933)
(D,0.7776)
};
\addlegendentry{Static}
\addplot [fill=blue,opacity=1.00]
coordinates {
(C,0.6357)
(A,0.7667)
(B,0.8326)
(D,0.8070)
};
\addlegendentry{Periodic}
\addplot [fill=red,opacity=1.00]
coordinates {
(C,0.6221)
(A,0.7601)
(B,0.8314)
(D,0.8000)
};
\addlegendentry{Drift-aware}
\end{axis}
\end{tikzpicture}}
    \\
    \multicolumn{2}{c}{\adjustbox{scale=0.85}{(e) Agent}}
    \end{tabular}
    \caption{Average results per family}\label{fig:final_res_model_per_family}
\end{figure}

\subsection{Accuracy per Training Period}

\begin{figure}[!htb]
    \centering
    \begin{tabular}{cc}
    \adjustbox{scale=0.7}{%
    \includegraphics[scale=0.175]{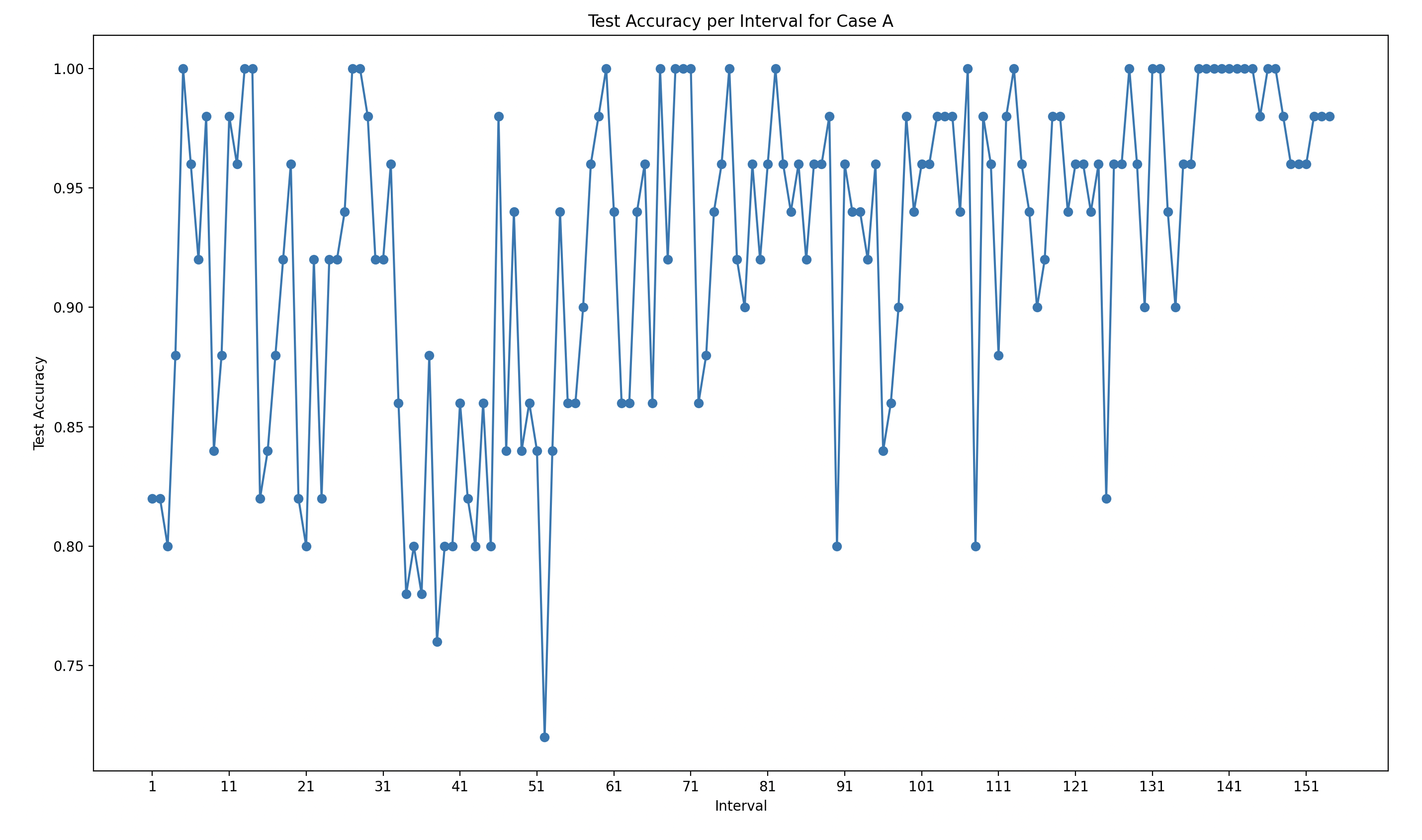}
    }
    &
    \adjustbox{scale=0.7}{%
    \includegraphics[scale=0.175]{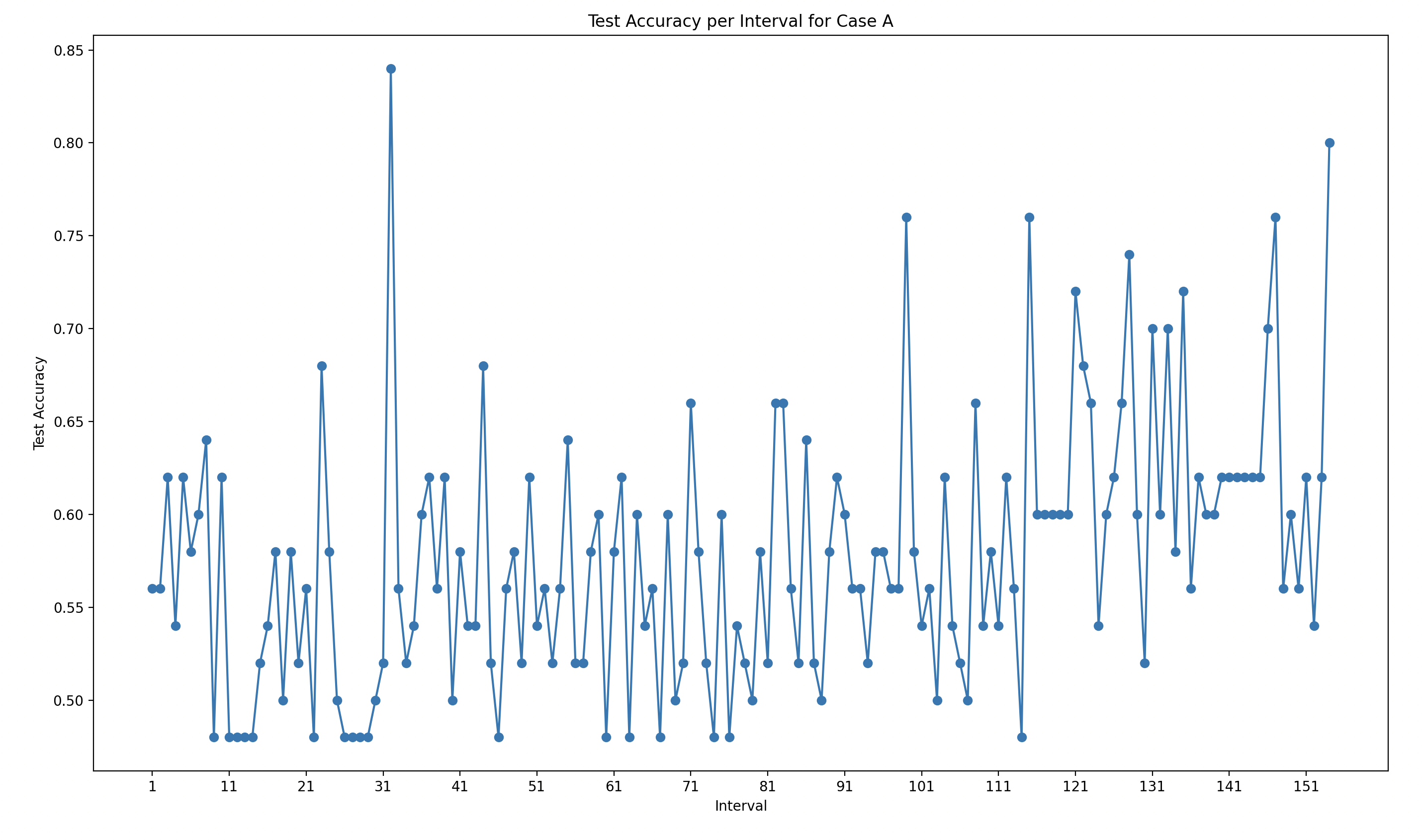}
    }
    \\
    \adjustbox{scale=0.85}{(a) Airpush vs Malap (XGBoost)}
    &
    \adjustbox{scale=0.85}{(b) Airpush vs SMSreg (MLP)}
    \\ \\[-0.5ex]
    \adjustbox{scale=0.7}{%
    \includegraphics[scale=0.175]{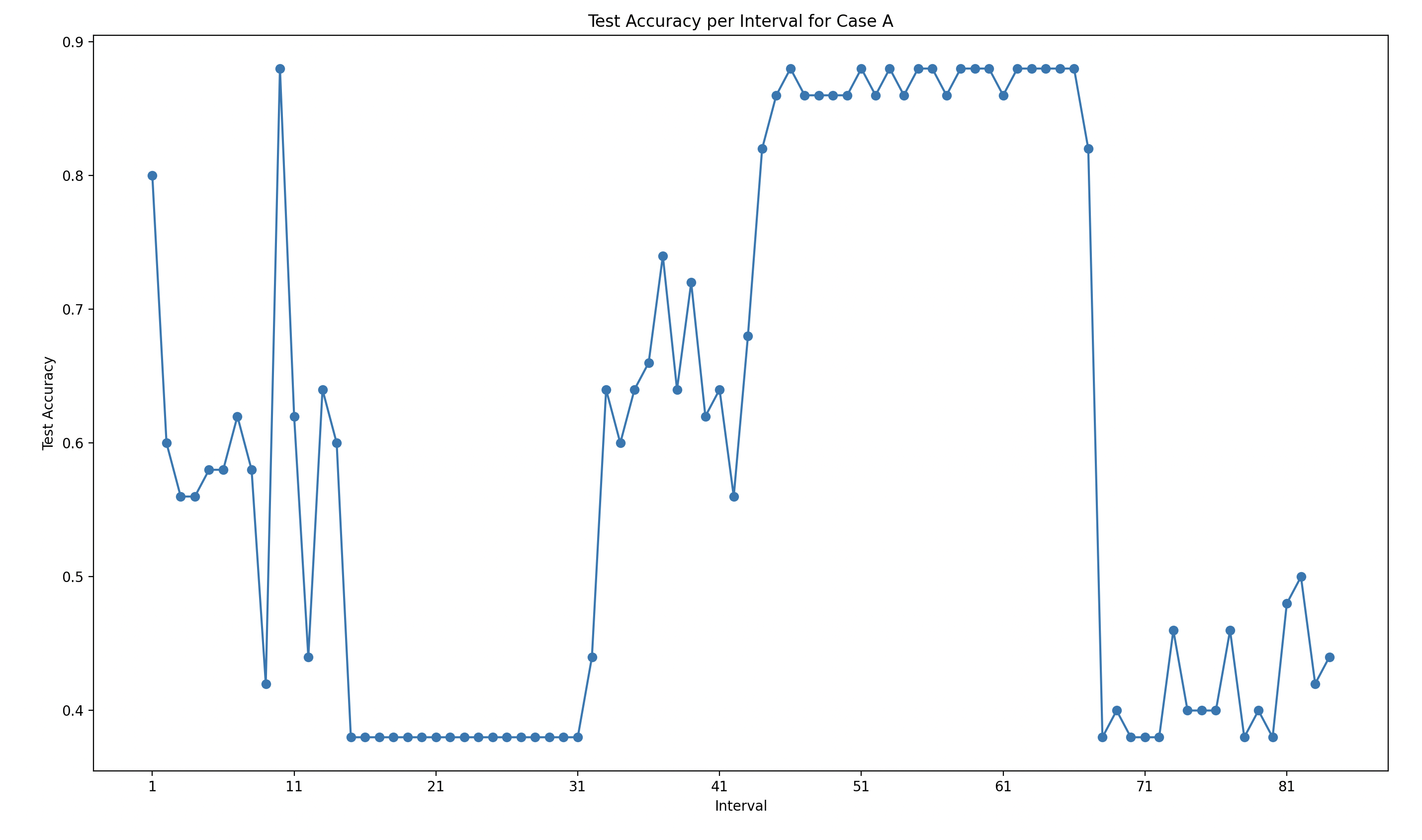}
    }
    &
    \adjustbox{scale=0.7}{%
    \includegraphics[scale=0.175]{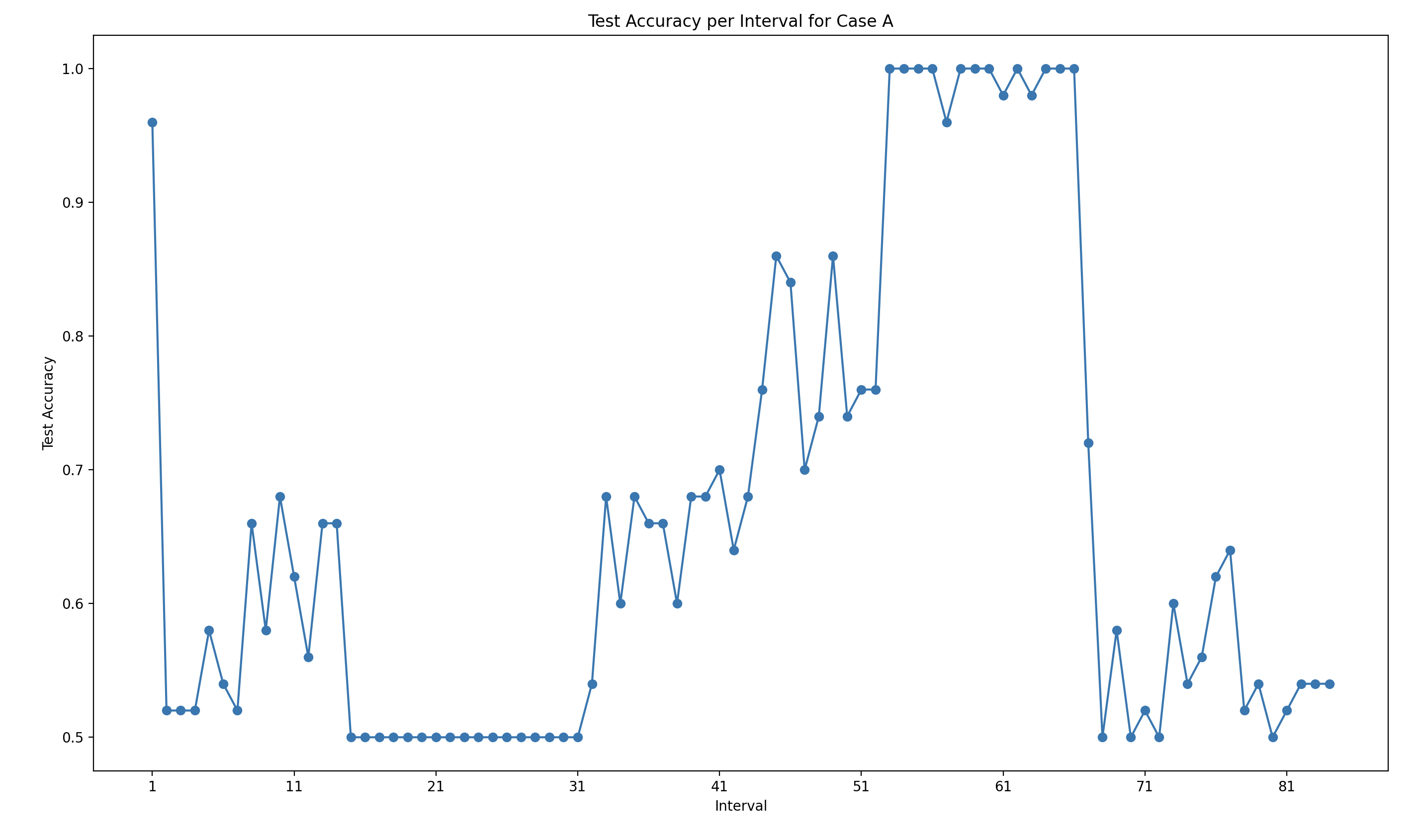}
    }
    \\
    \adjustbox{scale=0.85}{(c) SMSreg vs Agent (Random Forest)}
    &
    \adjustbox{scale=0.85}{(d) SMSreg vs Agent (SVM)}
    \end{tabular}
    \caption{Static training accuracy per interval (selected cases)}
    \label{fig:per_period_A}
\end{figure}

\begin{figure}[!htb]
    \centering
    \begin{tabular}{cc}
    \adjustbox{scale=0.7}{%
    \includegraphics[scale=0.175]{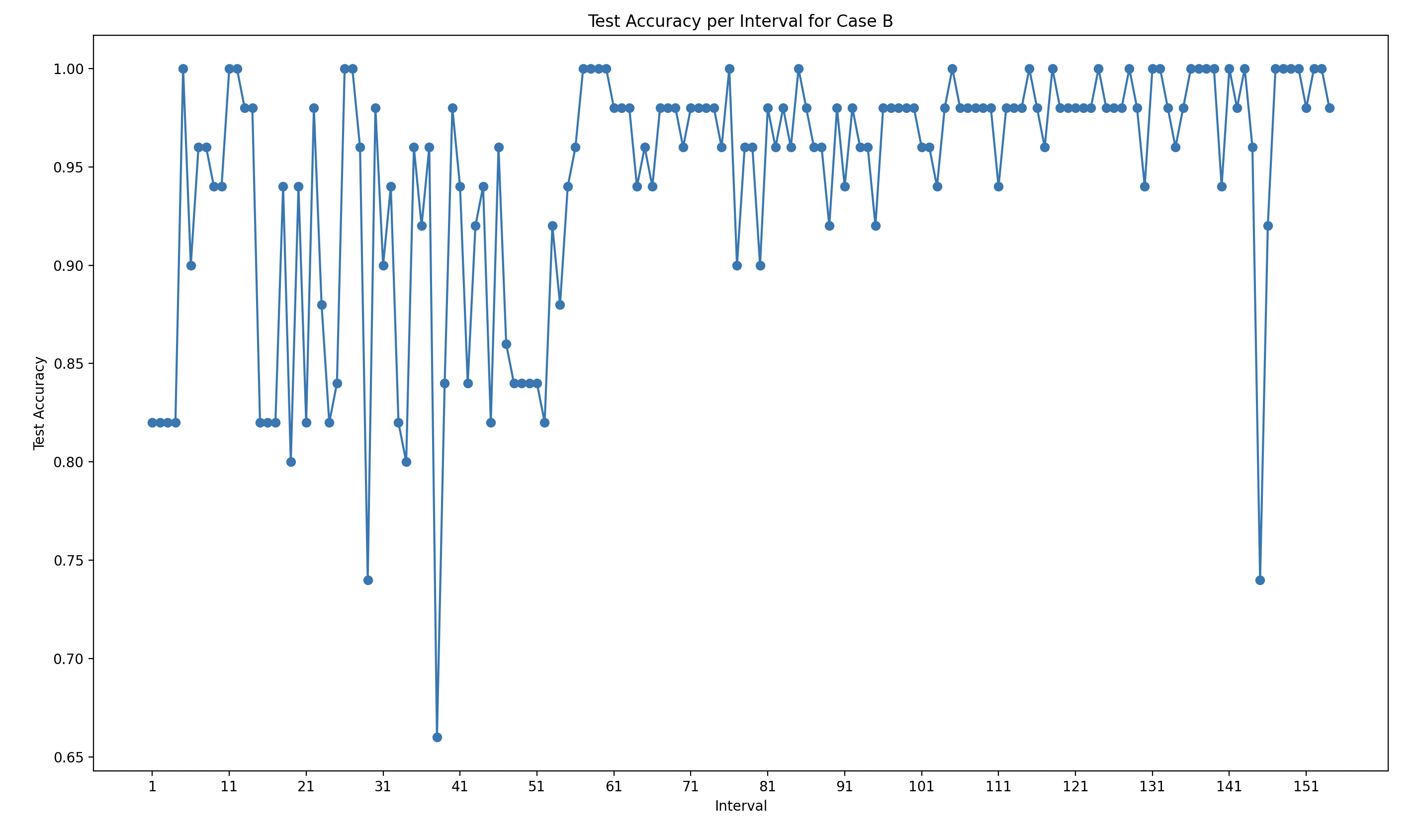}
    }
    &
    \adjustbox{scale=0.7}{%
    \includegraphics[scale=0.175]{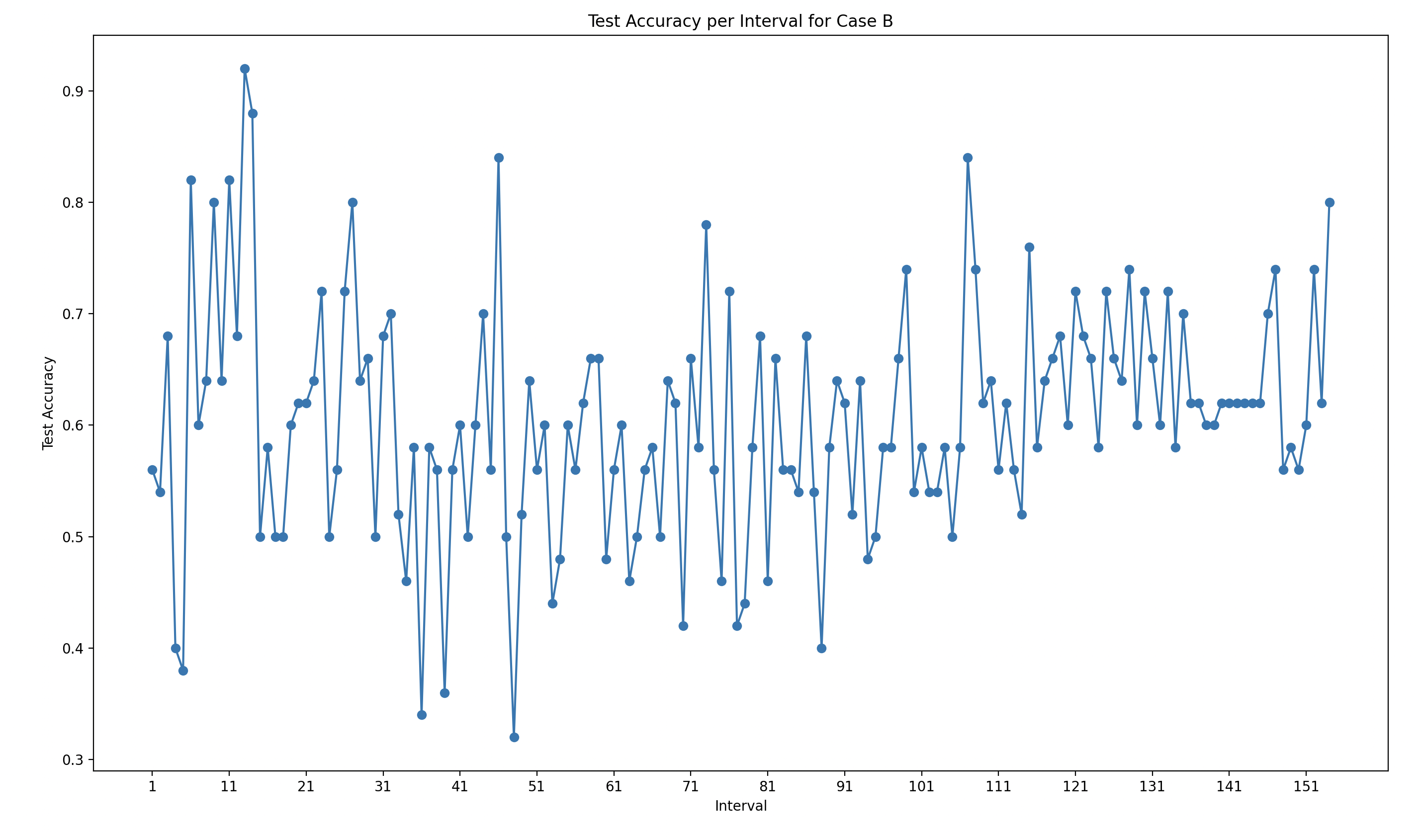}
    }
    \\
    \adjustbox{scale=0.85}{(a) Airpush vs Malap (XGBoost)}
    &
    \adjustbox{scale=0.85}{(b) Airpush vs SMSreg (MLP)}
    \\ \\[-0.5ex]
    \adjustbox{scale=0.7}{%
    \includegraphics[scale=0.175]{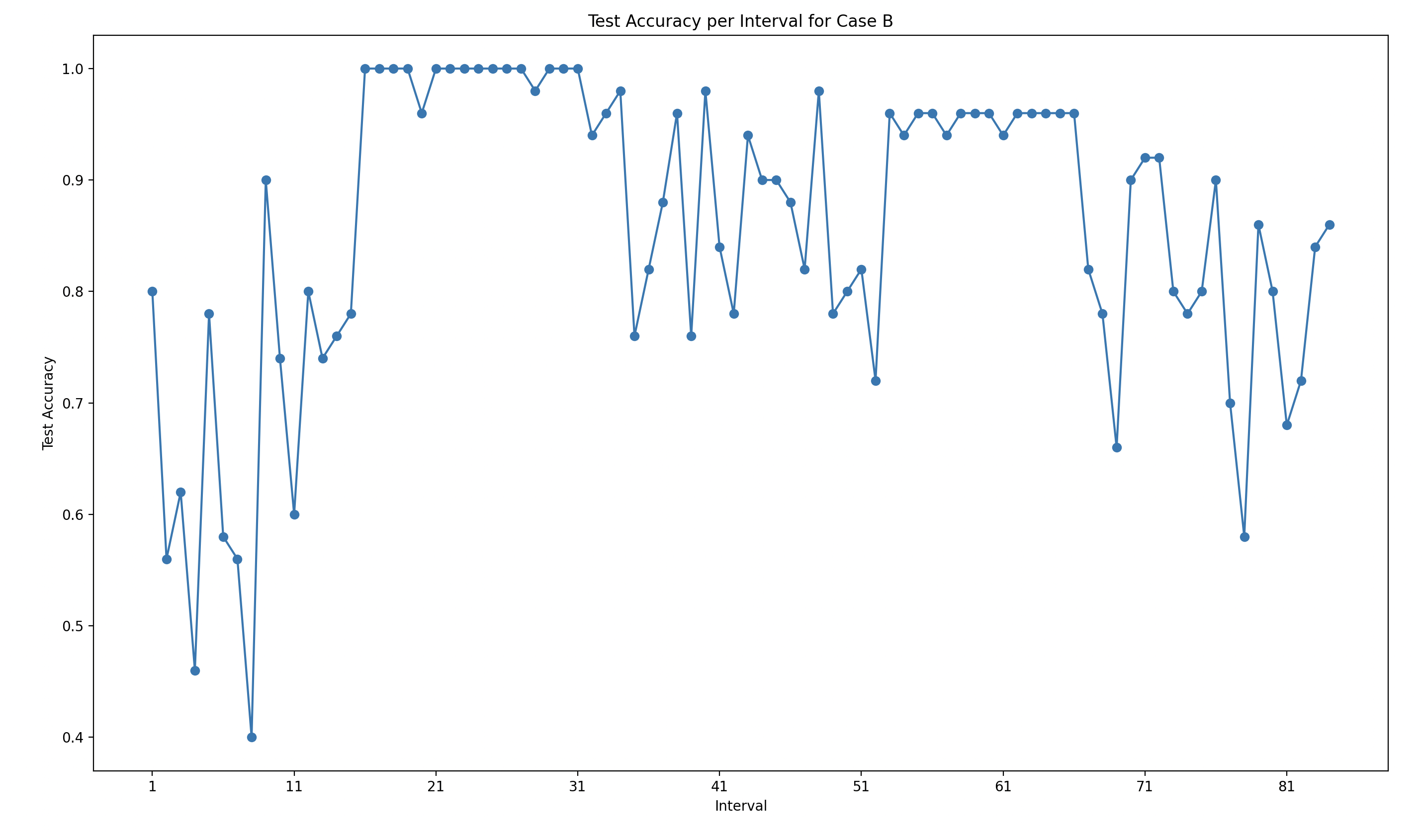}
    }
    &
    \adjustbox{scale=0.7}{%
    \includegraphics[scale=0.175]{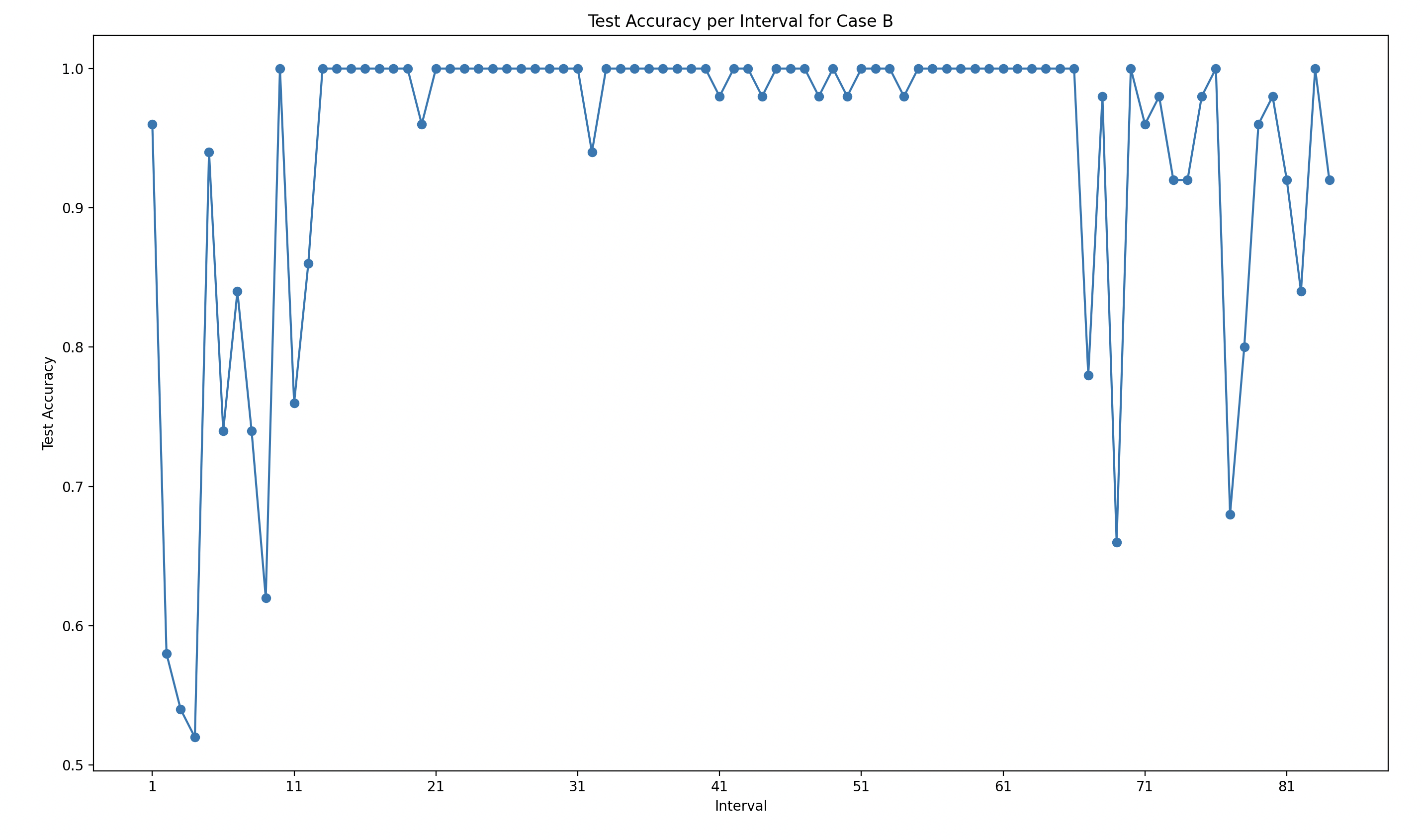}
    }
    \\
    \adjustbox{scale=0.85}{(c) SMSreg vs Agent (Random Forest)}
    &
    \adjustbox{scale=0.85}{(d) SMSreg vs Agent (SVM)}
    \end{tabular}
    \caption{Periodic retraining accuracy per interval (selected cases)}
    \label{fig:per_period_B}
\end{figure}

\begin{figure}[!htb]
    \centering
    \begin{tabular}{cc}
    \adjustbox{scale=0.7}{%
    \includegraphics[scale=0.175]{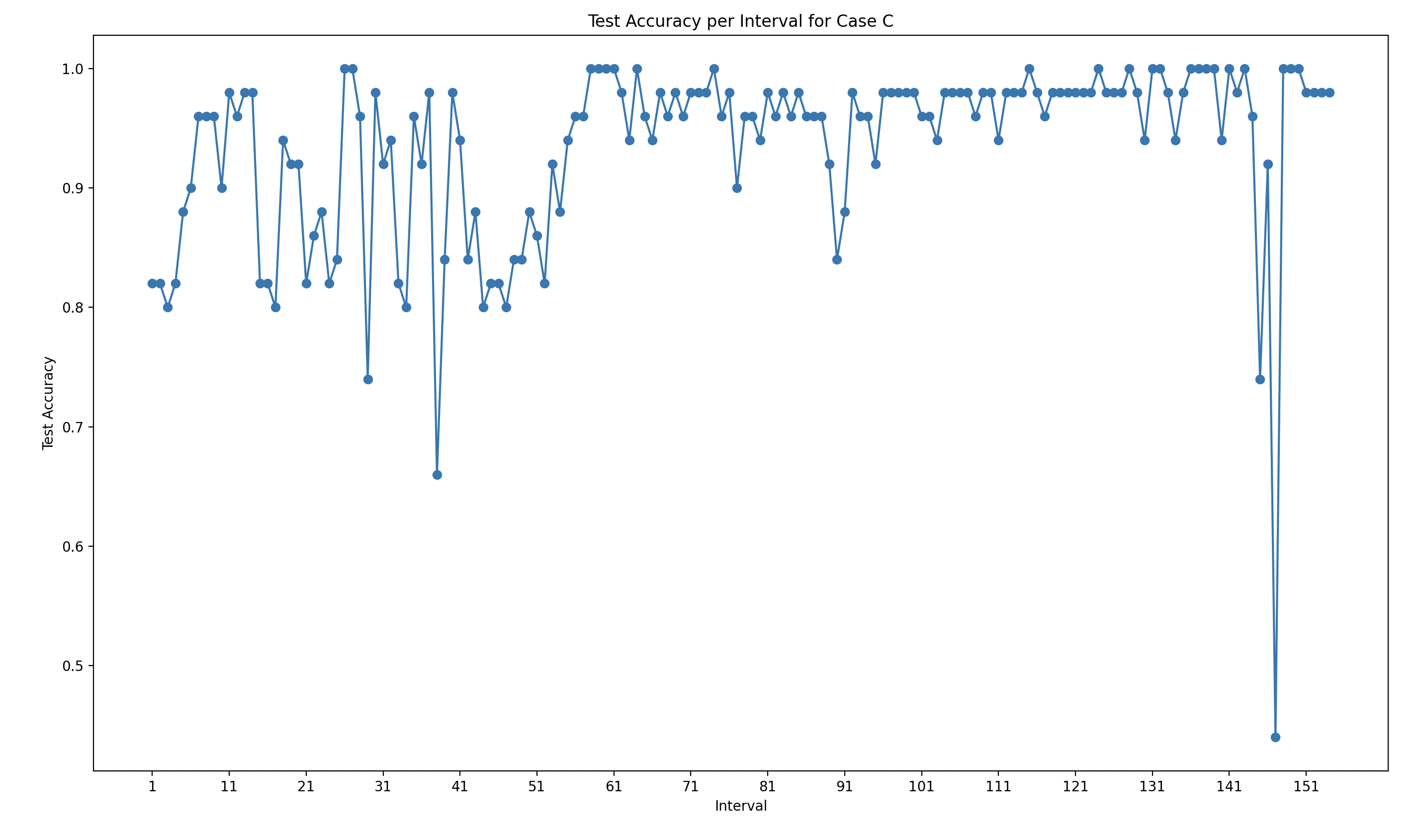}
    }
    &
    \adjustbox{scale=0.7}{%
    \includegraphics[scale=0.175]{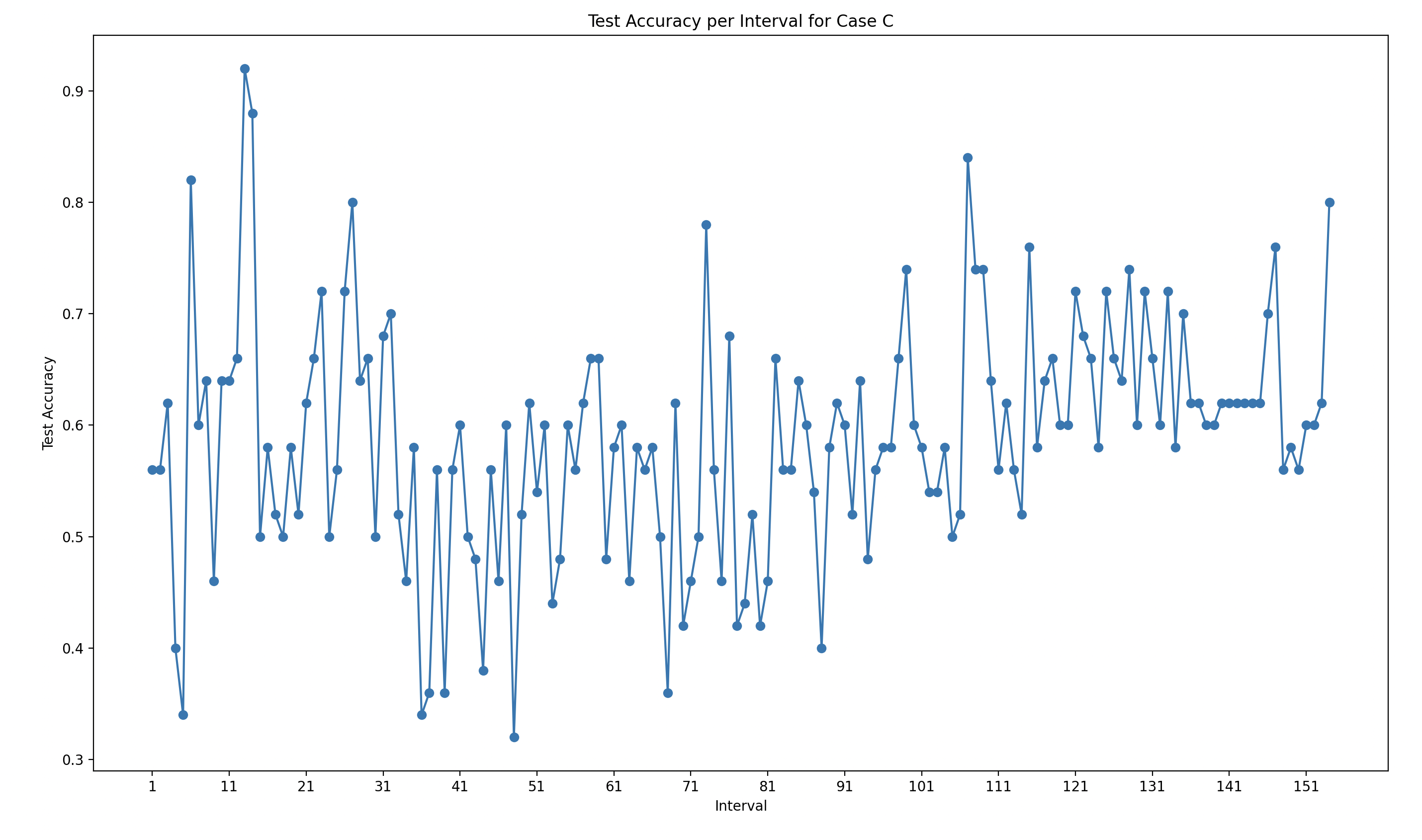}
    }
    \\
    \adjustbox{scale=0.85}{(a) Airpush vs Malap (XGBoost)}
    &
    \adjustbox{scale=0.85}{(b) Airpush vs SMSreg (MLP)}
    \\ \\[-0.5ex]
    \adjustbox{scale=0.7}{%
    \includegraphics[scale=0.175]{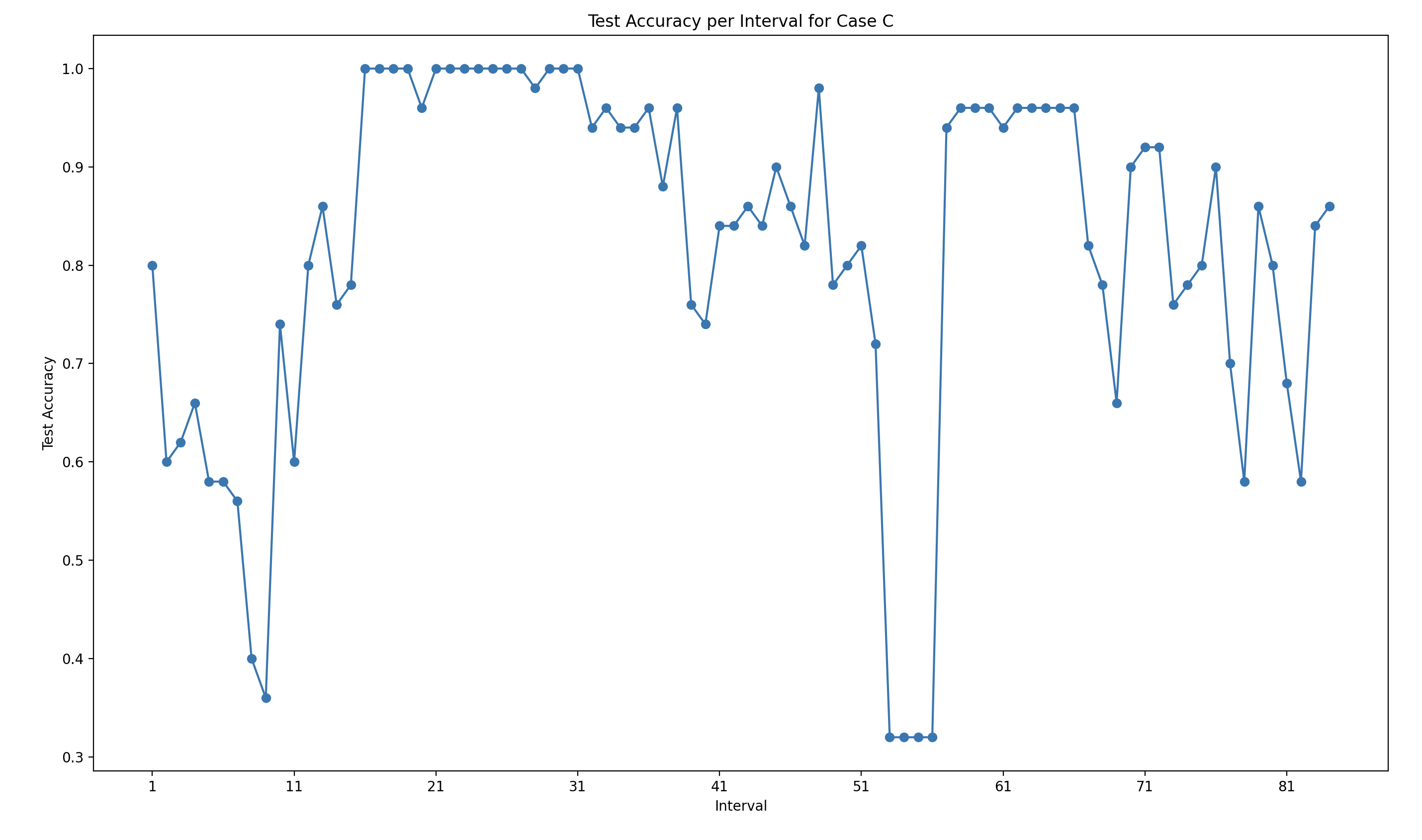}
    }
    &
    \adjustbox{scale=0.7}{%
    \includegraphics[scale=0.175]{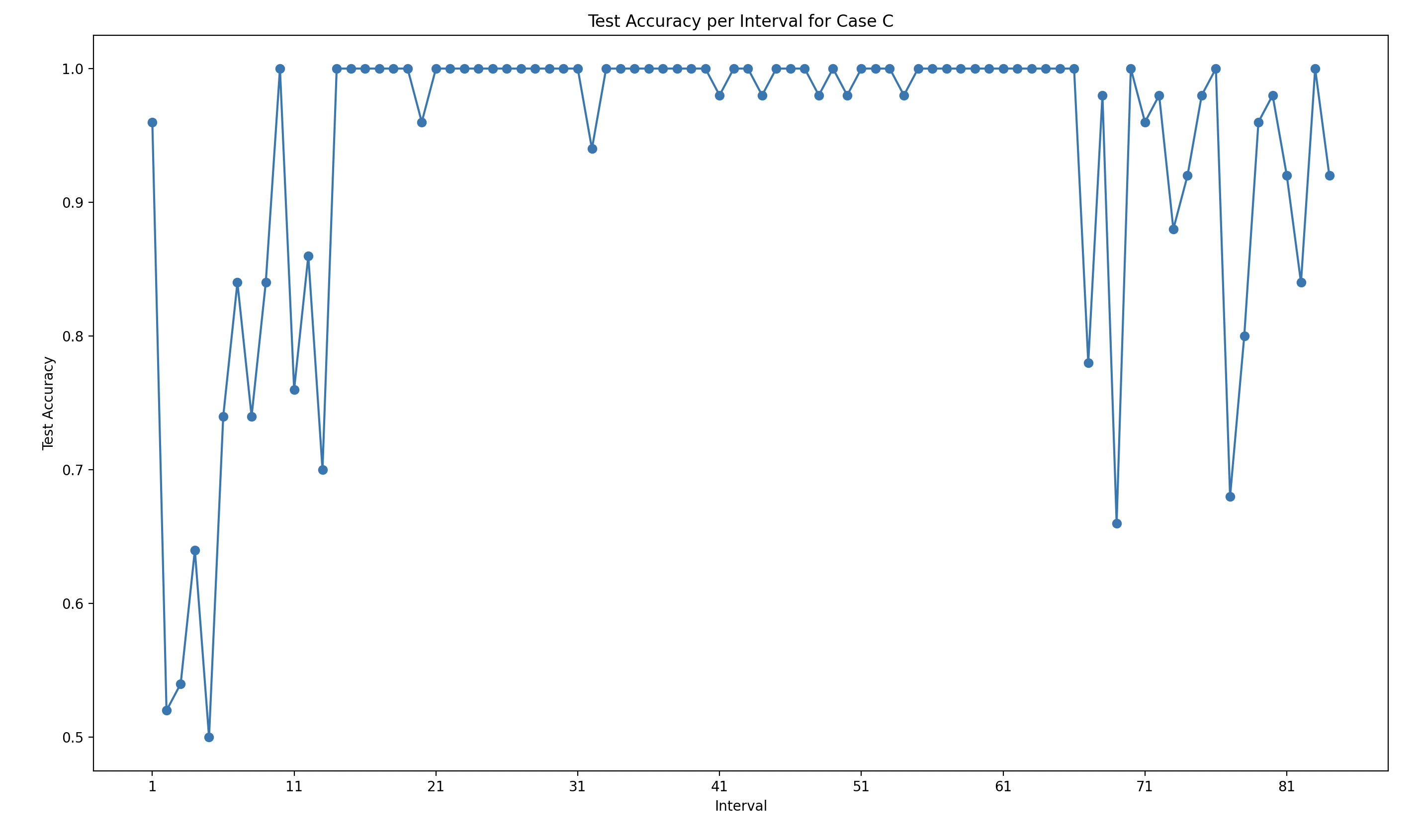}
    }
    \\
    \adjustbox{scale=0.85}{(c) SMSreg vs Agent (Random Forest)}
    &
    \adjustbox{scale=0.85}{(d) SMSreg vs Agent (SVM)}
    \end{tabular}
    \caption{Drift-aware retraining accuracy per interval (selected cases)}
    \label{fig:per_period_C}
\end{figure}

In this section, we provide graphs of model accuracy per training period 
for selected families 
and models. In Figure~\ref{fig:per_period_A} we give such results
for the static training scenario, while 
Figures~\ref{fig:per_period_B} and~\ref{fig:per_period_C} are
the corresponding results for the periodic retraining and drift-aware retraining scenarios,
respectively. These results are discuss in Section~\ref{sect:disc}.

\clearpage

\end{document}